\documentclass[final]{cvpr}
\usepackage{times}
\usepackage{epsfig}
\usepackage{graphicx}
\usepackage{amsmath}
\usepackage{amssymb}
\usepackage[linesnumbered,ruled,vlined]{algorithm2e}
\usepackage{verbatim} 
\usepackage{setspace}
\usepackage{bm}
\usepackage{mathrsfs}
\usepackage{multirow} 
\usepackage{makecell}

\usepackage{caption}
\usepackage{subcaption}
\usepackage{array}
\makeatletter
\newcommand{\thickhline}{%
    \noalign {\ifnum 0=`}\fi \hrule height 1pt
    \futurelet \reserved@a \@xhline
}

\usepackage[pagebackref=true,breaklinks=true,colorlinks,bookmarks=false]{hyperref}

\def\cvprPaperID{1162} 
\def\confYear{CVPR 2021}

\begin{document}

\title{Dive into Ambiguity: Latent Distribution Mining and Pairwise Uncertainty Estimation for Facial Expression Recognition}

\author{
Jiahui She$^{*1,2}$ \quad Yibo Hu$^{*2}$ \quad Hailin Shi$^{2}$
\quad Jun Wang$^{2}$ \quad Qiu Shen$^{\dag1}$ \quad Tao Mei$^{2}$\\
$^{1}$Nanjing University \quad
$^{2}$JD AI Research \quad \\
{\tt\small jh.she@foxmail.com, huyibo871079699@gmail.com, \{wangjun492, shihailin\}@jd.com} \quad \\
{\tt\small shenqiu@nju.edu.cn, tmei@live.com}
}

\maketitle
\pagestyle{empty}  
\thispagestyle{empty} 

\begin{abstract}
\vspace{-0.2cm}
Due to the subjective annotation and the inherent inter-class similarity of facial expressions, one of key challenges in Facial Expression Recognition (FER) is the annotation ambiguity. In this paper, we proposes a solution, named DMUE, to address the problem of annotation ambiguity from two perspectives: the latent \textbf{D}istribution \textbf{M}ining and the pairwise \textbf{U}ncertainty \textbf{E}stimation. For the former, an auxiliary multi-branch learning framework is introduced to better mine and describe the latent distribution in the label space. For the latter, the pairwise relationship of semantic feature between instances are fully exploited to estimate the ambiguity extent in the instance space. The proposed method is independent to the backbone architectures, and brings no extra burden for inference. The experiments are conducted on the popular real-world benchmarks and the synthetic noisy datasets.
Either way, the proposed DMUE stably achieves leading performance.
\end{abstract}

\let\thefootnote\relax\footnotetext{\vspace{-1pt}$*$ These authors contributed equally to this work.}
\let\thefootnote\relax\footnotetext{$\dag$ Corresponding author}

\vspace{-0.2cm}
\section{Introduction}\label{section:sec1}
Facial expression plays an essential role in human's daily life. Automatic Facial Expression Recognition (FER) is crucial in real world applications, such as service robots, driver fragile detection and human computer interaction. In recent years, with the emerge of large-scale datasets, \eg AffectNet~\cite{DBLP:journals/affectnet}, RAF-DB~\cite{DBLP:conf/cvpr/raf} and EmotioNet~\cite{DBLP:conf/cvpr/emotionet}, many deep learning based FER approaches~\cite{DBLP:conf/cvpr/LDL-ALSG, DBLP:conf/cvpr/SCN, DBLP:conf/eccv/IPA2LT} have been proposed and achieved promising performance.

\begin{figure}[t]
    \centering
    \includegraphics[width=0.96\linewidth]{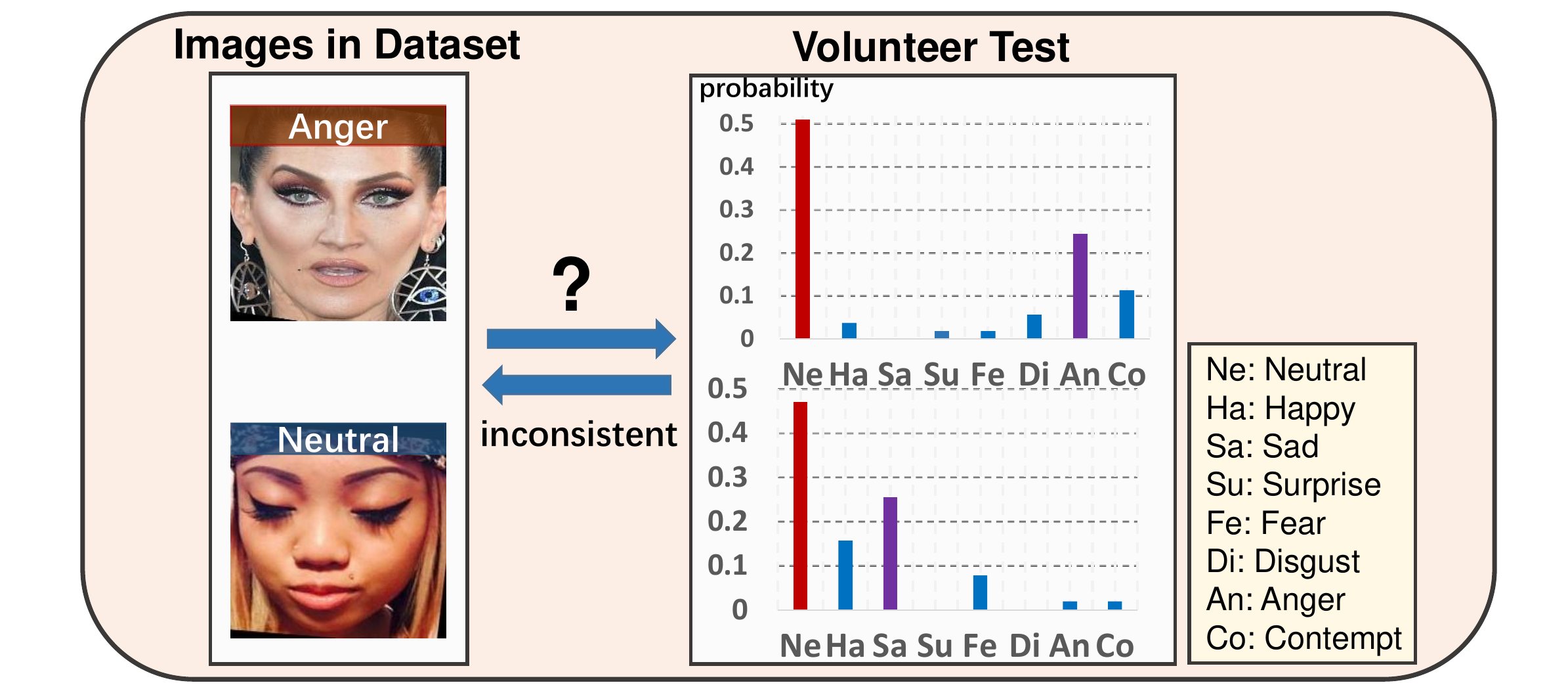}
    \vspace{-0.2cm}
    \caption{User study results by 50 volunteers on two randomly picked images.
    The red (purple) bar represents the most (secondary) possible class given by the volunteers.
    The results provide insights that the annotations may be inconsistent among the users.}
    \vspace{-0.2cm}
    \label{fig:demo}
\end{figure}

However, the ambiguity problem remains an obstacle that hinders the FER performance. Usually, facial images are annotated to one of several basic expressions for training the FER model.
Yet the definition with respect to the expression category may be inconsistent among different people. For better understanding, we randomly pick two images from AffectNet~\cite{DBLP:journals/affectnet} and conduct a user study. As shown in Fig.~\ref{fig:demo}, for the image annotated with \textit{Anger}, the most possible class decided by volunteers is \textit{Neutral}. For the other image, the confidence gap between the most and secondary possible classes is only 20\%, which means annotating it to a specific class is not suitable. In other words, a label distribution that depicts the possibility belonging to each class can better describe the visual feature. There are two reasons leading to the above phenomenon: (1) It is subjective for people to define which type of expression a facial image is. (2) With a large amount of images in large-scale FER datasets, it is expensive and time-consuming to provide label distribution of images. As there exists a considerable portion of ambiguous samples in large-scale datasets, the models are prevented from learning the robust visual features with respect to a certain type of expression, thus the performance has reached a bottleneck. The previous approaches tried to address this issue by introducing label distribution learning~\cite{DBLP:conf/cvpr/LDL-ALSG} or suppressing uncertain samples~\cite{DBLP:conf/cvpr/SCN}. However, they still suffer from the ambiguity problem revealed in data that cannot be directly solved from the single instance perspective.

In this paper, we propose a solution to address the ambiguity problem in FER from two perspectives, \ie the latent \textbf{D}istribution \textbf{M}ining and the pairwise \textbf{U}ncertainty \textbf{E}stimation (DMUE).
For the former, several temporary auxiliary branches are introduced to discover the label distributions of samples in an online manner. The iteratively updated distributions can better describe the visual features of expression images in the label space. Thus, it can provide the model informative semantic features to flexibly handle ambiguous images. For the latter, we design an elaborate uncertainty estimation module based on pairwise relationships between samples. It jointly utilizes the original annotations and the statistics of relationships to reflect the ambiguity extent of samples. The estimated uncertain level encourages the model to dynamically adjust learning focus between the mined label distribution and original annotations. Note that our proposed framework is end-to-end training and has no extra cost for inference. All the auxiliary branches and the uncertainty estimation module will be removed during deployment. Overall, the main contributions can be summarized as follows:

\begin{itemize}
\item We propose a novel end-to-end solution to investigate the ambiguity problem in FER by exploring the latent label distribution of the given sample, without introducing extra burden on inference.

\item An elaborate uncertainty estimation module is designed based on the statistics of relationships, which provides guidance for the model to dynamically adjust learning focus between the mined label distribution and annotations from sample level.

\item Our approach is evaluated on the popular real-world benchmarks and synthetic noisy datasets. Particularly, it achieves the best performance by 89.42\% on RAF-DB and 63.11\% on AffectNet, setting new records.
\end{itemize}

\begin{figure*}
    \centering
    \includegraphics[width=0.86\linewidth]{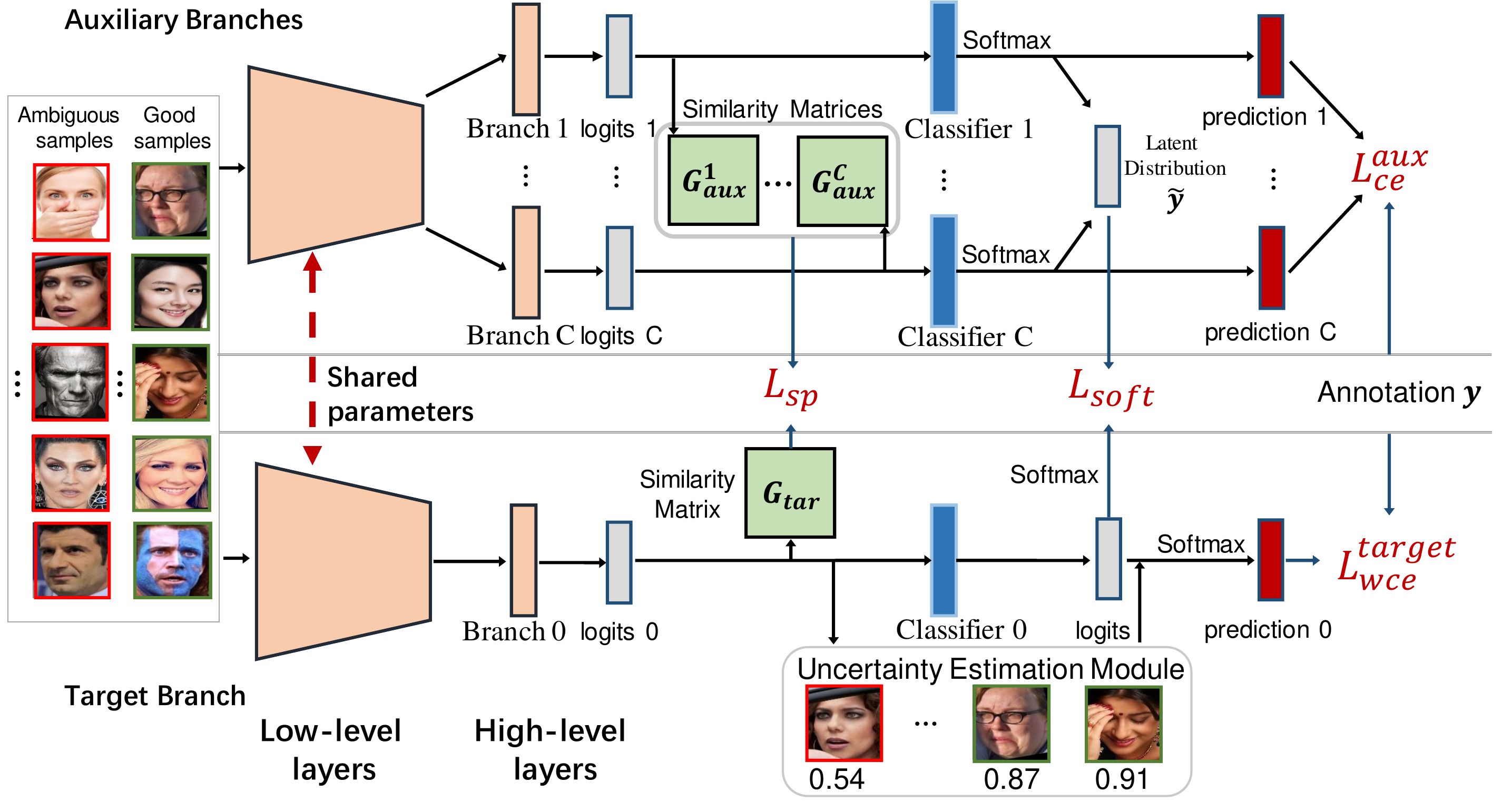}
    \vspace{-1em}
    \caption{Overview of the DMUE. $\boldsymbol{y}$ denotes the set of annotations of images in a batch. $\boldsymbol{\widetilde{y}}$ denotes the set of mined latent distributions of images in a batch.}
    \label{fig:method}
    \vspace{-1em}
\end{figure*}

\vspace{-0.1cm}
\section{Related Work}
\subsection{Facial Expression Recognition}
Numerous FER algorithms~\cite{SeNet50,DBLP:journals/tip/gaCNN,DBLP:journals/SIFT,DBLP:conf/mm/normal_acm1} have been proposed, which can be grouped into \textit{handcraft and learning-based} methods. Early attempts~\cite{DBLP:conf/fgr/Gabor_Wavelets, DBLP:journals/SIFT, DBLP:journals/LBP} rely on handcraft features that reflect folds and geometry changes caused by expression. With the development of deep learning, \textit{learning-based} methods~\cite{DBLP:journals/corr/R3A1, DBLP:conf/mm/normal_acm1,DBLP:conf/cvpr/DeRl} become the majority, such as decoupling the identity information~\cite{DBLP:conf/mm/normal_acm1} or exploiting the difference between expressive images~\cite{DBLP:conf/cvpr/DeRl}.

In recent years, several attempts try to address ambiguity problem in FER. Zeng \etal~\cite{DBLP:conf/eccv/IPA2LT} consider annotation inconsistency and introduce multiple training phases. Chen \etal~\cite{DBLP:conf/cvpr/LDL-ALSG} build nearest neighbor graphs for training data in advance and investigate label distribution of samples in a semi-online way. Previous leading performance has been achieved by Wang \etal~\cite{DBLP:conf/cvpr/SCN}. They focus on finding the confidence weight and the latent truth of each sample to suppress harmful influence from ambiguous data. However, the compound expressions~\cite{DBLP:conf/cvpr/raf} and the original annotations could be jointly considered in estimating ambiguity.

\subsection{Learning with Ambiguity Label} 
Mislabelled annotations and low data quality may result in ambiguity problem. For the former, learning with noisy label~\cite{DBLP:journals/corr/noisy_label_survey} is one of the most popular directions. Another direction is the uncertainty estimation~\cite{DBLP:conf/nips/metaweightnet, DBLP:conf/iccv/noisy_2}, such as MentorNet~\cite{DBLP:conf/icml/MentorNet} and CleanLab~\cite{DBLP:journals/corr/cleanlab}. In recent years, a promising way to handle mislabelled annotations is to find the latent truth~\cite{DBLP:conf/iclr/noiselayer}, such as utilizing the prediction of model~\cite{DBLP:conf/nips/MixMatch,DBLP:conf/iclr/noiselayer, DBLP:conf/iccv/label_noisy1_phototype,DBLP:conf/iclr/dividemix,DBLP:journals/ijcv/AberHu2} or introducing auxiliary embeddings~\cite{DBLP:conf/eccv/AberHu1,DBLP:conf/cvpr/label_noisy1}. For the latter, an universal way is to enhance the label~\cite{genxin1, genxin2} of low quality images by the temperature softmax~\cite{DBLP:conf/nips/MixMatch} or inject the artificial uncertainty~\cite{DBLP:conf/cvpr/propface2, DBLP:conf/iccv/propface, DBLP:conf/cvpr/uncertainty1}. Unlike prior methods~\cite{genxin1, genxin2}, the ambiguity problem in FER, \ie compound expressions~\cite{DBLP:conf/cvpr/raf} exists in a more subjective way. The label description of a compound expression image is various among the users.


\section{Method}
\vspace{-0.1cm}
\textbf{Notation.} Given a FER dataset $(\mathcal{X}, \mathcal{Y})$ in which each image $\boldsymbol{x}$ belongs to one of $C$ classes, we denote $y_x\in\{1,2,\cdots,C\}$ as its annotated deterministic class. However, as shown in Fig.~\ref{fig:demo}, the exact type of $\boldsymbol{x}$ is inapparent or uncertain. We employ \textit{latent distribution} $\widetilde{\boldsymbol{y}}_x$ to represent the probability distribution for $\boldsymbol{x}$ belonging to all possible classes except $y_x$. That is, $\widetilde{\boldsymbol{y}}_x\in\mathbb{R}^{C-1}$ is a distribution vector, $\| \widetilde{\boldsymbol{y}}_x \|_1=1$.

\subsection{Overview of DMUE}
\vspace{-0.06cm}
To address the annotation ambiguity, we mine $\widetilde{\boldsymbol{y}}_x$ for each $\boldsymbol{x}$ and regularize the model to learn jointly from $\widetilde{\boldsymbol{y}}_x$ and $y_x$. Benefited from the semantic features of ambiguous samples, the performance of the model can be greatly improved.

For a toy experiment, we train a ResNet-18 on AffectNet~\cite{DBLP:journals/affectnet} and present its prediction for a mislabelled training sample in Fig.~\ref{fig:toy}, where a crying baby (\textit{Sad}) is labelled with \textit{Neutral}. We can observe that the predicted distribution reflects the truth class of the mislabelled sample. It inspires us to employ the predictions from a trained model to help a new model in training phase, where such mislabelled image may be tagged by a distribution reflecting its true class. By imposing the latent distribution $\widetilde{\boldsymbol{y}}_x$ as the additional supervision, model can utilize the latent semantic features to better deal with ambiguity samples, thus improve the performance. We employ the classifier trained by samples from negative classes of $\boldsymbol{x}$, \ie samples from other $C-1$ classes except for $y_x$, to find its $\widetilde{\boldsymbol{y}}_x$, based on qualitative and quantitative analyses in Section~\ref{sec:ablation}. Moreover, to balance the learning between the annotation and the mined $\widetilde{\boldsymbol{y}}_x$, an uncertainty estimation module is elaborately designed to guide the model to learn more from $\widetilde{\boldsymbol{y}}_x$ than $y_x$ for those ambiguous samples.

An overview of DMUE is depicted in Fig.~\ref{fig:method}. The DMUE contains: (1) latent distribution mining with C auxiliary branches and one target branch that have the same architecture (\eg the last stage of ResNet), and (2) pairwise uncertainty estimation, where an uncertainty estimation module is established by two fully connected (FC) layers. Each auxiliary branch is served as an individual $(C-1)$-class classifier aiming to find $\widetilde{\boldsymbol{y}}_x$ for the corresponding $\boldsymbol{x}$. $\widetilde{\boldsymbol{y}}_x$ and $y_x$ are joint together to guide the target branch. Furthermore, we regularize the branches to predict consistent relationships of images by their similarity matrices. Note that all auxiliary branches and the uncertainty estimation module will be removed, and only the target branch will be reserved for deployment. \textit{Therefore, our framework is end-to-end and can be flexibly integrated into existing network architectures without extra cost on inference.}

\subsection{Latent Distribution Mining}
As $\widetilde{\boldsymbol{y}}_x$ is predicted by the classifier trained with samples from negative classes of $\boldsymbol{x}$, If there are total $C$ classes, then $C$ classifiers need to be trained to predict the latent distribution of each sample. Considering the computational efficiency and the shared low-level features~\cite{DBLP:conf/aaai/online3,DBLP:conf/nips/online2}, we propose a multi-branch architecture to construct these classifiers. As shown in Fig.~\ref{fig:method}, $C$ auxiliary branches are introduced to predict the latent distributions and a target branch is employed for final prediction. Given a batch, the $j$-th branch predicts $\widetilde{\boldsymbol{y}}_x$ for $\boldsymbol{x}$ annotated to the $j$-th class. Thus, we can obtain $\widetilde{\boldsymbol{y}}_x$ for each $\boldsymbol{x}$ in batch by $C$ auxiliary branches. Note all branches have the same structure (\eg the last stage of ResNet) and share the common lower layers (\eg the first three stages of ResNet). Classifier $j,j\in\{1,\cdots,C\}$ is $(C-1)$-class and Classifier $0$ (the target classifier) is $C$-class for final deployment.

A comprehensive description of mini-batch training is presented in Algorithm~\ref{alg:algorithm1}. Given a batch, we use images not annotated to the $j$-th category to train the $j$-th auxiliary branch. In other words, each image $\boldsymbol{x}$ is utilized to train other $C-1$ auxiliary branches than the $y_x$-th branch. The Cross-Entropy(CE) loss $L_{CE}^{aux}$ is employed for optimization:
\par\noindent
\begin{equation}
    \vspace{-1em}
    \centering
    {\rm{L}}_{CE}^{aux} = \frac{1}{C}\sum\limits_{j = 1}^C {L_{CE}^{au{x_j}},}
    \vspace{-0.5em}
\end{equation}
\begin{equation}
    \centering
    {\rm{L}}_{CE}^{au{x_j}} =  - \frac{\begin{array}{l}
\\
1
\end{array}}{N_j}\sum\limits_{p = 1}^{N_j} {\sum\limits_{k = 1, k \ne j}^{C} {{y_{x_p,k}}\log {f_j}({\boldsymbol{x}_p};\theta )_k} ,}
\end{equation}\label{eq:trainaux}
\noindent
where $L_{CE}^{aux_j}$ is the CE loss for training the $j$-th branch, $N_j$ is the number of $\boldsymbol{x}$ not annotated to $j$ in the batch and $p$ is index. $y_{x_p,k}$ is the label of  $\boldsymbol{x}_p$ belonging to the $k$-th class and ${f_j}{({\boldsymbol{x}_p};\theta)_k}$ is the possibility of $\boldsymbol{x}_p$ belonging to the $k$-th class predicted by the $j$-th branch.

As described above, the prediction of the $j$-th auxiliary branch for $\boldsymbol{x}$ with annotation $j$, is used as its latent distribution $\widetilde{\boldsymbol{y}}_x\in\mathbb{R}^{C-1}$. One additional step, called Sharpen~\cite{DBLP:conf/nips/MixMatch, DBLP:conf/iclr/dividemix, DBLP:journals/corr/online1}, is adopted before regularizing the target branch:
\par\noindent
\begin{equation}
    \centering
    Sharpen(\widetilde{\boldsymbol{y}}_x, T)_i=\widetilde{y}_{x,i}^{\frac{1}{T}}/\sum\nolimits_j^{C - 1} {\widetilde{y}_{x,j}^{\frac{1}{T}},}
\end{equation}
\par\noindent
where $\widetilde{{y}}_{x,i}$ the $i$-th element of $\widetilde{\boldsymbol{y}}_x$ and $T$ is the temperature. Sharpen function provides the flexibility to slightly adjust the entropy of $\widetilde{\boldsymbol{y}}_x$. When $T>1$, the output $Sharpen(\widetilde{\boldsymbol{y}}_x,T)$ will be more flatten than the original $\widetilde{\boldsymbol{y}}_x$.

After sharpening, we utilize $L_2$ loss to minimize the deviation between the prediction of target branch and the sharpened $\widetilde{\boldsymbol{y}}_x$, which is defined as:
\noindent
\begin{equation}
    \centering
    {L_{soft}} = \frac{1}{{N(C - 1)}}\sum\limits_{p = 1}^N {\sum\limits_{k = 1,k \ne i}^C {{{({\widetilde{y}_{{x_p},k}} - {f_{target}}{{({\boldsymbol{x}_p};\theta )}_k})}^2}}, }
\end{equation}\label{eq:L2loss}
\noindent
where $N$ is the batch size. $\widetilde{y}_{{x_p},k}$ is the possibility of $\boldsymbol{x}_p$ belonging to the $k$-th class in the latent distribution $\widetilde{\boldsymbol{y}}_{x_p}$, and ${f_{target}}{{({\boldsymbol{x}_k};\theta )}_j}$ is the prediction of target branch. The reason employing L2 loss is that unlike Cross-Entropy, the L2 loss is bounded and less sensitive to inaccurate predictions. We do not back propagate gradients through computing $\widetilde{\boldsymbol{y}}$.

\textbf{Similarity Preserving.} Inspired by ~\cite{DBLP:conf/iccv/SP-KD}, we find it is beneficial to regularize all the branches to predict consistent relationship when given a pair of images. This is because CE loss only utilizes samples individually from the label space. However, the relationship between samples is another knowledge paradigm. For instance, given a pair of \textit{smiling} images, besides telling network their annotations \textit{Happy}, the similarities of their semantic features extracted by different branches should be consistent. Thus, we generalize ~\cite{DBLP:conf/iccv/SP-KD} to the context of multi-branch architecture as multi-branch similarity preserving ($MSP$), defined as:
\begin{equation}
    \centering
    {L_{sp}} = MSP(G_{aux}^1,\cdots,G_{aux}^C, G_{tar}),
\end{equation}
where $G_{aux}^i\in\mathbb{R}^{B_i\times{B_i}}$ and $G_{tar}\in\mathbb{R}^{B\times{B}}$ are the similarity matrices calculated by the semantic features in auxiliary and target branches, respectively. Their elements reflect the pairwise relationships between samples. $L_{sp}$ aims at sharing the relation information across branches. \textit{Specific computation is provided in the supplementary material.}

\begin{figure}[t]
    \centering
    \includegraphics[width=0.95\linewidth]{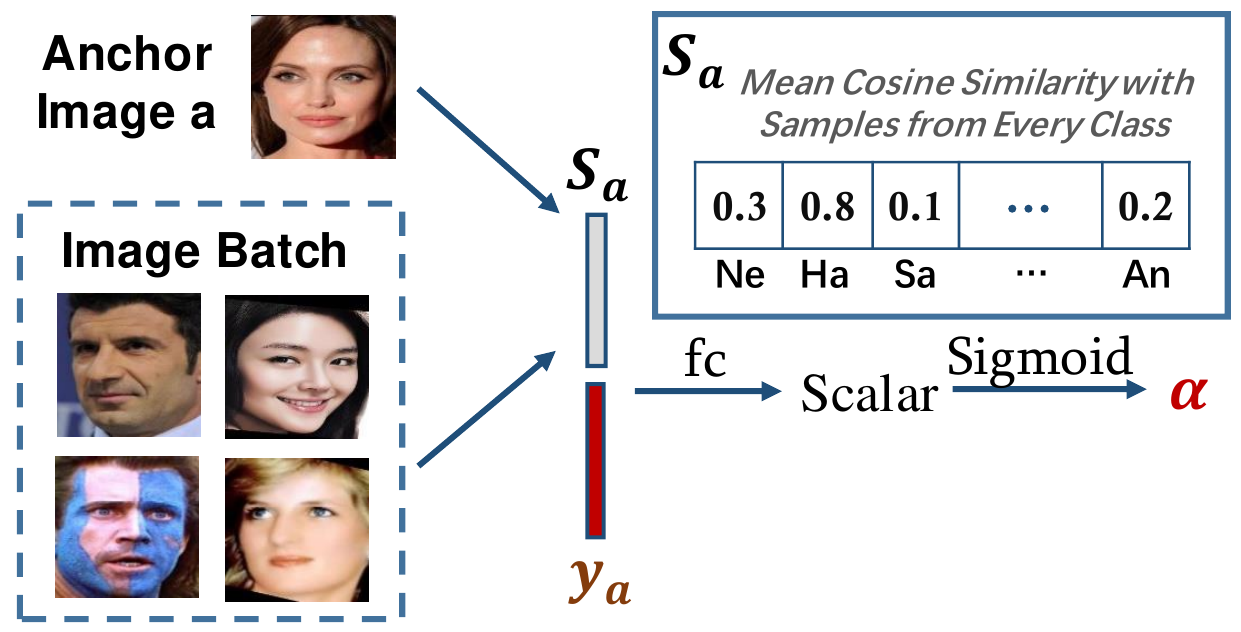}
    \vspace{-0.3cm}
    \caption{Uncertainty estimation module. $y_a$ is the one-hot form of anchor image's annotation. $S_a$ and $y_a$ are concatenated to reflect how ambiguous the anchor image is. }
    \label{fig:estimation}
    \vspace{-0.3cm}
\end{figure}
\subsection{Pairwise Uncertainty Estimation}
To handle the ambiguous samples, we introduce latent distribution mining. However, the target branch should also be benefitted from clean samples. Directly employing CE loss may lead to improvement degradation due to the existing of ambiguous samples. Accordingly, we impose a modulator term into the standard CE loss to trade-off between the latent distribution and annotation in the sample space. Specifically, we estimate the confidence scores of the samples based on the statistics of their relationships. Lower score will be assigned to more ambiguous samples, further reducing the CE loss. Thus, the latent distribution will provide more guidance.

For better understanding, we choose an anchor image in a given batch to illustrate the uncertainty estimation module. As shown in Fig.~\ref{fig:estimation}, we denote the semantic feature and one-hot label of the anchor image as $(\boldsymbol{f}_a, \boldsymbol{y}_a)$, while others in the batch as $(\boldsymbol{f}_i, \boldsymbol{y}_i)$, where $\boldsymbol{f}$ is the feature before classifier in the target branch, $i$ is the index. We calculate the average cosine similarity of $\boldsymbol{f}_a$ with each of $\boldsymbol{f}_i$ annotated with $j$-th category as $S_{a,j}$, and vector $\boldsymbol{S}_a = [S_{a,1},\cdots,S_{a,C}]$. After that, $\boldsymbol{S}_a$ is concatenated with $\boldsymbol{y}_a\in\mathbb{R}^{C}$ to form $\boldsymbol{SV}_a\in\mathbb{R}^{2C}$, which reflects how ambiguous the anchor sample is:
\begin{equation}
    \centering
    \boldsymbol{SV}_a = concat(\boldsymbol{S}_a,\boldsymbol{y}_a),
\end{equation}
\par\noindent
\begin{equation}
    \centering
    {S_{a,j}} = \frac{1}{{{N_j}}}\sum\limits_{i = 1}^{{N_j}} {\frac{{\left< {{\boldsymbol{f}_a},{\boldsymbol{f}_i}} \right>}}{{\left\| {{\boldsymbol{f}_a}} \right\|\left\| {{\boldsymbol{f}_i}} \right\|}},}
\end{equation}
where $<\boldsymbol{f}_a,\boldsymbol{f}_i>$ is the dot product of $\boldsymbol{f}_a$ and $\boldsymbol{f}_i$. $N_j$ is the number of samples whose annotation is $j$-th class in the batch and $i$ is the index.

Here, We provide two perspectives to understand this delicate design: (\textbf{1}) For a mislabelled sample $(\boldsymbol{x},y_x)$ in the given batch (\eg the semantic feature of $\boldsymbol{x}$ belongs to $i$-th class but $y_x=j$), the average similarity of semantic features between $\boldsymbol{x}$ and the images in $i$-th class should be high. However, the concatenated $y_x$ indicates $\boldsymbol{x}$ is annotated with $i$-th class. (\textbf{2}) A \textit{clear} $\boldsymbol{x} (y_x=i)$ should only capture the typical semantic feature of $i$-th class. The average similarity of its semantic feature with other types of images should be discriminatively lower than with $i$-th class samples. Thus, $\boldsymbol{SV}_x$ can reveal the ambiguity information of $\boldsymbol{x}$.

Let $\boldsymbol{SV}=[\boldsymbol{SV}_1, \boldsymbol{SV}_2,\cdots,\boldsymbol{SV}_N]\in\mathbb{R}^{2C\times{N}}$ denotes the ambiguity information feature of a batch, the uncertainty estimation module takes $\boldsymbol{SV}$ as the input and outputs a confidence scalar $\alpha_i\in(0,1)$ for each image. The module consists of two FC layers with a PRelu non-linear function and a sigmoid activation:
\begin{equation}
    \centering
    {\alpha} = Sigmoid(\boldsymbol{W}_2^T\sigma (\boldsymbol{W}_1^T\boldsymbol{SV}),
\end{equation}
where $\boldsymbol{W}_1\in\mathbb{R}^{2C\times{C}}$ and $\boldsymbol{W}_2\in\mathbb{R}^{C\times{1}}$ are the parameters of two FC layers,$\sigma$ is the PReLU activation.

With the estimated confidence score, we perform weighted training in the target branch. Directly multiplying the score with CE loss may obstruct the uncertainty estimation, because it will make the estimated score to be zero~\cite{DBLP:conf/cvpr/SCN}. Therefore, we alternatively multiply the score with the output logit of the classifier in the target branch. The weighted CE loss~\cite{DBLP:conf/cvpr/face1,DBLP:conf/cvpr/SCN} is formulated as:
\begin{equation}
    \centering
    L_{WCE}^{t\arg et} =  - \frac{1}{N}\sum\limits_{i = 1}^N {\log \frac{{{e^{{\alpha _i}W_{{y_i}}^T{f_i}}}}}{{\sum\nolimits_{j = 1}^C {{e^{{\alpha _i}W_j^T{f_i}}}} }}.}
\end{equation}\label{eq:target}
\noindent
Obviously, $L_{WCE}^{target}$ has positive correlation with the score $\alpha$~\cite{DBLP:conf/cvpr/Sphereface}. Thus, for ambiguous samples, the estimated scores are small, reducing the impact of CE loss, and the target branch learns more from the mined latent distributions.
\subsection{Overall Loss function}
The overall objective of DMUE is:
\noindent
\begin{equation}
    \centering
    L_{total} = w_{u}(e)(L_{WCE}^{target} + \omega{L_{soft}} + \gamma{L_{sp}}) + w_{d}(e){L_{CE}^{aux}},
\end{equation}
\noindent
where $\omega$, $\gamma$ are the hyperparameters. $w_{u}$ and $w_{d}$ are the weighted ramp functions~\cite{DBLP:conf/iclr/ramp} \wrt the epoch \textit{e}, which is formulated as:
\noindent
\begin{equation}
\label{rampup}
w_{u}(e) = \left\{ {\begin{array}{*{20}{c}}
{\exp ( -{{(1 - \frac{e}{\beta })}^2})}&{e \le \beta }\\
1&{e > \beta }
\end{array}} \right.,
\end{equation}
\begin{equation}
    \label{rampdown}
w_{d}(e) = \left\{ {\begin{array}{*{20}{c}}
1&{e \le \beta }\\
{\exp ( -{{(1 - \frac{\beta }{e})}^2})}&{e > \beta }
\end{array}} \right.,
\end{equation}
where $\beta$ is the epoch threshold for functions.
\noindent
where $\beta$ is the epoch threshold. The Eq.~\ref{rampup} and~\ref{rampdown} are introduced to benefit training from two aspects: (1) At the beginning of training, the latent distributions mined by auxiliary branches are not stable enough. Thus, we focus on training the auxiliary branches. (2) When the auxiliary branches are well trained, we then divert our attention to train the target branch.

It worth noting that we remove all the auxiliary branches and the uncertainty estimation module for deployment. \textit{Our framework is end-to-end and can be flexibly integrated with existing network architectures, without extra cost on inference.}

\section{Experiments}
We verify the effectiveness of DMUE on synthetic noisy datasets and 4 popular in-the-wild benchmarks, and further validate the contribution of each component of DMUE. Extensive ablation studies with respect to the hyperparameters and the different backbone architectures are carried out to confirm the advantage of our method.
\setlength{\textfloatsep}{10pt}
\begin{algorithm}[tb]
    \SetNoFillComment 
    \SetCommentSty{mycommfont} 
	\caption{DMUE.} 
	\label{alg:algorithm1}
     \KwIn{Training Images $\mathcal{X}$ and annotations $\mathcal{Y}$ with $C$ classes, $MaxEpoch$, $num\_iters$}
     \KwOut{Trained model with target branch $\theta^{0}$ and $C$ auxiliary branches $\theta^{j},j\in\{1,\cdots, C\}$\\}
    \tcc{\textit{Training}}
	Initialize $\theta^{0}$ and $\theta^{j}$ with random values, $j\in\{1,2,\cdots, C\}, e=1$\\
	\While{\textnormal{$e < MaxEpoch$}}{
	      \For{$k=0$ \KwTo num\_iters}{
	        From $(\mathcal{X}, \mathcal{Y})$, sample a batch $set_{batch}$\tcp*{Note samples in $j$-th class as $set_j$}
            Compute $\mathcal{L}_{ce}^{aux}$
            \tcp*{use $set_{batch}\backslash set_j$ to compute $\mathcal{L}_{ce}^{aux_j}$ for $\theta^j,j\in\{1,\cdots, C\}$}

            Compute \textit{latent distribution} for $set_{batch}$
              \tcp*{Use $\theta^j$ predict for $set_j,$ $j\in\{1,\cdots, C\}$}

            Compute $\mathcal{L}_{soft}$ and $\mathcal{L}_{sp}$

            Compute $\mathcal{L}_{wce}^{target}$ in $\theta^{0}$
            \tcp*{use $set_{batch}$}

            Update all branches $\theta^{j},j\in\{0,1,2,\cdots, C\}$
            }
        $e = e + 1$;
	}
	\tcc{\textit{Testing}}
	Deploy model only with the target branch $\theta^{0}$
\end{algorithm}

\subsection{Datasets and Metrics}\label{sec:dataset}
\textbf{RAF-DB}~\cite{DBLP:conf/cvpr/raf} is constructed by 30,000 facial images with basic or compound annotations. In the experiment, we choose the images with seven basic expressions (\ie neutral, happiness, surprise, sadness, anger, disgust and fear), of which 12,271 are used for training, and the remaining 3,068 for testing.
\textbf{AffectNet}~\cite{DBLP:journals/affectnet} is currently the largest FER dataset, including 440,000 images. The images are collected from the Internet by querying the major search engines with 1,250 emotion-related keywords. Half of the images are annotated with eight basic expressions, providing 280K training images and 4K testing images.
\textbf{FERPlus}~\cite{DBLP:conf/icmi/ferplus} is an extension of FER2013~\cite{DBLP:conf/iconip/fer2013}, including 28,709 training images and 3,589 testing images resized to 48$\times$48 gray-scale pixels. Each image is labelled by 10 crowd-sourced annotators to one of eight categories. For a fair comparison, the most voting category is picked as the annotation for each image following ~\cite{DBLP:conf/icmi/ferplus,resvgg, DBLP:conf/cvpr/SCN,DBLP:journals/tip/RAN}.
\textbf{SFEW}~\cite{DBLP:conf/sfew} contains the images from movies with seven basic emotions, including 958 images for training and 436 images for testing. For each dataset, we report the overall accuracy on the testing set.
\subsection{Implementation Details}\label{}
By default, we use ResNet-18 as the backbone network pretrained on MS-Celeb-1M with the standard routine~\cite{DBLP:conf/cvpr/SCN, DBLP:journals/tip/RAN} for a fair comparison. The last stage and the classifier of ResNet-18 are separated tor form auxiliary branches, while the remaining low-level layers are shared across auxiliary and target branches. The facial images are aligned and cropped with three landmarks~\cite{DBLP:conf/iccv/adaptiveWing}, resized to 256$\times$256 pixels, augmented by random cropping to 224$\times$224 pixels and horizontal flipped with a probability of $0.5$. During training, the batch size is 72, and each batch is constructed to ensure every class is included. We use Adam with weight decay of $10^{-4}$. The initial learning rate is $10^{-3}$, which is further divided by 10 at epoch 10 and 20. The training ends at epoch 40. Only the target branch is kept during testing. By default, the hyperparameters are set as $T=1.2, \omega=0.5, \beta=6$ and $\gamma=10^{3}$, according to the ablation studies.
All experiments are carried out on a single Nvidia Tesla P40 GPU which takes 12 hours to train AffectNet with 40 epochs.

\subsection{Evaluation on Synthetic Ambiguity} \label{sec:synthetic}
The annotation ambiguity in FER mainly lies in two aspects: mislabelled annotations and uncertain visual representation. We quantitatively evaluate the improvement of DMUE against the mislabelled annotations on RAF-DB and AffectNet.
Specifically, a portion (\eg 10\%, 20\% and 30\%) of the training samples are randomly chosen, of which the labels are flipped to other random categories.
We choose ResNet-18 as the baseline and the backbone of DMUE, and compare the performance with SCN~\cite{DBLP:conf/cvpr/SCN}, which is the state-of-the-art noise-tolerant FER method. SCN reckons uncertainty in each sample by its visual feature, and aims to find their deterministic latent truth. Each experiment is repeated three times, then the mean accuracy and standard deviation on the testing set are reported. To make fair comparison, SCN is pretrained on MS-Celeb-1M with the backbone of ResNet-18.

As shown in Table~\ref{Noisy}, the DMUE outperforms each baseline and SCN~\cite{DBLP:conf/cvpr/SCN} consistently. With noise ratio of 30\%, DMUE improves the accuracy by 4.29\% and 4.21\% on RAF-DB and AffectNet, respectively. This attributes to the mined latent distribution that can flexibly describe both synthetic noisy samples and compound expressions in the label space. Thus, it guides the model to overcome the harmful influence from noisy annotations.

\textbf{Visualization of $\widetilde{\boldsymbol{y}}$.}
Qualitative results are presented in the supplementary material to demonstrate that our approach can obtain the latent truth for mislabelled samples, and thereby achieve performance improvement.


\subsection{Component Analysis}
We conduct experiments on RAF-DB and AffectNet to analyse the contribution of latent distribution mining, uncertainty estimation and similarity preserving. As shown in Table~\ref{ComponentEvaluation}, some observations can be found: \textbf{(1)} Latent distribution mining plays a more important role than others. When only one component employed, it outperforms similarity preserving and uncertainty estimation by 2.09\% and 0.4\% on AffectNet, 1.19\% and 0.13\% on RAF-DB, respectively. It proves the benefits provided by the latent distribution, as the semantic features of ambiguous images are well utilized. \textbf{(2)} When combining uncertainty estimation and latent distribution, we achieve performance improvement by 0.74\% and 0.91\% over only using the latent distribution on AffectNet and RAF-DB, respectively. It attributes to the uncertainty estimation module providing guidance for the target branch. Thus, the target branch can flexibly adjust the learning focus between the annotation and the latent distribution, according to the ambiguous extent of samples. \textbf{(3)} Similarity preserving also brings some improvements, while its contribution is relatively small than others. As it benefits the learning mainly by making different branches predict consistent relationships for image pairs, speeding up the training convergence. \textit{We present more results of similarity preserving in the supplementary material.}

\begin{table}[t]
\begin{center}
\caption{Mean Accuracy and standard deviation ($\%$) on RAF-DB and AffectNet with synthetic noisy annotations.}\label{Noisy}
\vspace{-0.5em}
\resizebox{0.9\columnwidth}{!}{ 
\begin{tabular}{c|c|c|c}
\thickhline
Method & Noisy(\%) & RAF-DB & AffectNet\\
\hline\hline
Baseline & 10 & 80.43$\pm$0.72 & 57.21$\pm$0.31\\
SCN~\cite{DBLP:conf/cvpr/SCN} & 10 & 81.92$\pm$0.69 & 58.48$\pm$0.62\\
DMUE & 10 & \textbf{83.19$\pm$0.83} & \textbf{61.21$\pm$0.36} \\ \hline\hline
Baseline & 20 & 78.01$\pm$0.29 & 56.21$\pm$0.31\\
SCN~\cite{DBLP:conf/cvpr/SCN} & 20 & 80.02$\pm$0.32 & 56.98$\pm$0.28\\
DMUE & 20 & \textbf{81.02$\pm$0.69} & \textbf{59.06$\pm$0.34} \\ \hline\hline
Baseline & 30 & 75.12$\pm$0.78 & 52.67$\pm$0.45\\
SCN~\cite{DBLP:conf/cvpr/SCN} & 30 & 77.46$\pm$0.64 & 55.04$\pm$0.54\\
DMUE & 30 & \textbf{79.41$\pm$0.74} & \textbf{56.88$\pm$0.56} \\ \thickhline

\end{tabular}} %
\end{center}
\vspace{-1em}
\end{table}

\begin{table}[t]
\vspace{-0.5em}
\begin{center}
\caption{Accuracy ($\%$) comparison of the different components.
SP denotes the similarity preserving. Confidence denotes involving the uncertainty estimation module for the weighted training in target branch.}\label{ComponentEvaluation}
\vspace{-0.5em}
\resizebox{\columnwidth}{!}{ 
\begin{tabular}{c|c|c|c|c}
\thickhline
Latent distribution & SP & Confidence & AffectNet & RAF-DB\\
\hline\hline
- & - & - & 58.85 & 86.33\\
\checkmark & - & - & 61.76 & 87.84\\
- & \checkmark & - & 59.67 & 86.65\\
- & - & \checkmark & 61.36 & 87.71\\
\checkmark & \checkmark & - & 62.34 & 88.23\\
- & \checkmark & \checkmark & 61.65 & 87.98\\
\checkmark & - & \checkmark & 62.50 & 88.45\\
\hline\hline
\checkmark & \checkmark & \checkmark & \textbf{62.84} & \textbf{88.76}\\
\thickhline
\end{tabular}} %
\end{center}
\vspace{-1em}
\end{table}

\begin{figure*}[t]
    \centering
    \includegraphics[width=1\linewidth]{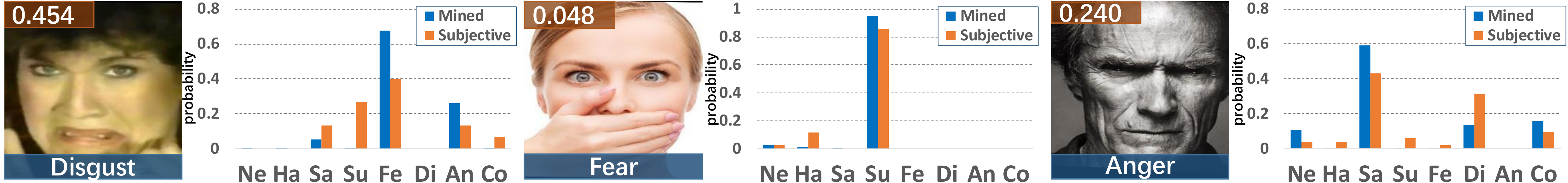}
    \vspace{-0.6cm}
    \caption{The mined latent distribution is compared with the subjective results. Each image is tagged with annotation and the KL-divergence between two distributions. The generated latent distribution is consistent with intuition. Best viewed in color. Zoom in for better view. (Ne=Neutral,Ha=Happy,Sa=Sad,Su=Surprise,Fe=Fear,Di=Disgust,An=Anger,Co=Contempt).
    }
    \label{fig:label_dis}
    \vspace{-0.3em}
\end{figure*}

\begin{figure}[t]
    \vspace{-1em}
    \centering
    \includegraphics[width=0.95\linewidth]{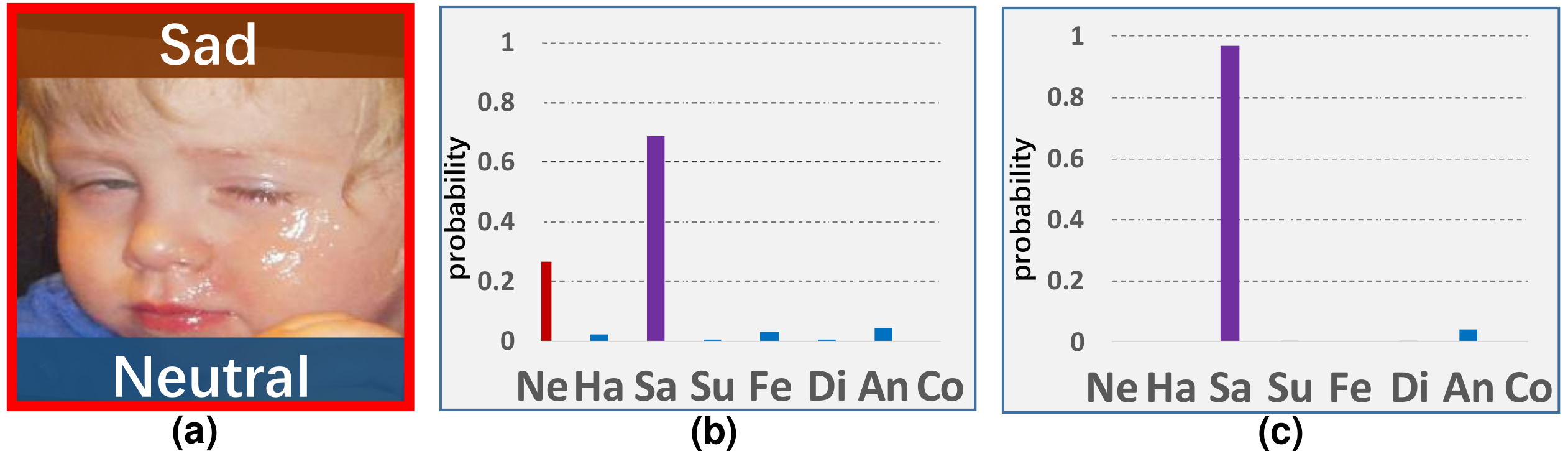}
    \vspace{-0.2cm}
    \caption{(a) A \textit{Sad} training image mislabelled with \textit{Neutral} in AffectNet. (b) The prediction from model trained on AffectNet can show the right class. (c) The prediction from model trained on all classes except for the \textit{Neutral} better reflects the truth discriminatively. Best viewed in color. Zoom in for better view.
    }
    \label{fig:toy}

\end{figure}

\subsection{Comparison with the State-of-the-art}
We compare DMUE with existing state-of-the-art methods on 4 popular in-the-wild benchmarks in Table~\ref{table:state-of-the-art}.

\textbf{Results.}
In Table~\ref{table:state-of-the-art}, both CAKE~\cite{DBLP:conf/bmvc/cake}, SCN~\cite{DBLP:conf/cvpr/SCN} and RAN~\cite{DBLP:journals/affectnet} utilize ResNet-18 as the backbone. SCN and RAN are pretrained on MS-Celeb-1M according to their original papers. RAN mainly deals with the occlusion and head pose problem in FER. As shown in Tabel~\ref{table:state-of-the-art}, DMUE achieves current leading performance on AffectNet. For RAF-DB, all three LDL-ALSG, IPA2LT and SCN are noise-tolerant FER methods considering ambiguity, among which SCN achieves state-of-the-art results. We further improve the performance of ambiguous FER by mining latent distribution and considering annotations in uncertainty estimation. Table~\ref{table:state-of-the-art} also shows the results on FERPlus and SFEW, respectively. Without bells and whistles, our method achieves better performance than the counterparts.

\subsection{Visualization Analysis}
To further diagnose our method, we conduct visualizations of the discovered latent distribution and the estimated confidence score.

\textbf{Latent Distribution.}
In Section~\ref{sec:synthetic}, we quantitatively demonstrate the effectiveness of DUME to deal with mislabelled images. In this section, we further conduct user study for qualitative analysis of how latent distribution cope with uncertain expressions. Specifically, 20 images are randomly picked from RAF-DB and AffectNet, and labelled by 50 voters. As latent distribution reflects the sample's probability distribution among its negative classes, we set the number of votes on sample's positive class to be zero. The normalized subjective results are compared with the mined latent distribution.

In Fig.~\ref{fig:label_dis}, KL-divergence between the subjective result and the latent distribution is reported for reference. It is interesting to see people have different views of the specific type of expressions. Furthermore, our approach obtains qualitatively consistent results with human intuition. Although there exists differences in details, it is worth noting that the results can already qualitatively explain that the latent distribution benefits the model by reinforcing the supervision information.

\textbf{Confidence Score.}
To corporate with latent distribution mining, a confidence score is estimated by the uncertainty estimation module for each image given a batch. The more ambiguous a sample is, the lower its confidence score will be. Thus, the target branch will learn more from its latent distribution. We qualitatively analyse the uncertainty estimation module by visualizing images with the original annotation and the scaled confidence score. Moreover, we rank images by their confidence scores and report their ranks in a batch of 72 images.

\begin{figure}[t]
    \centering
    \vspace{-1em}
    \includegraphics[width=0.95\linewidth]{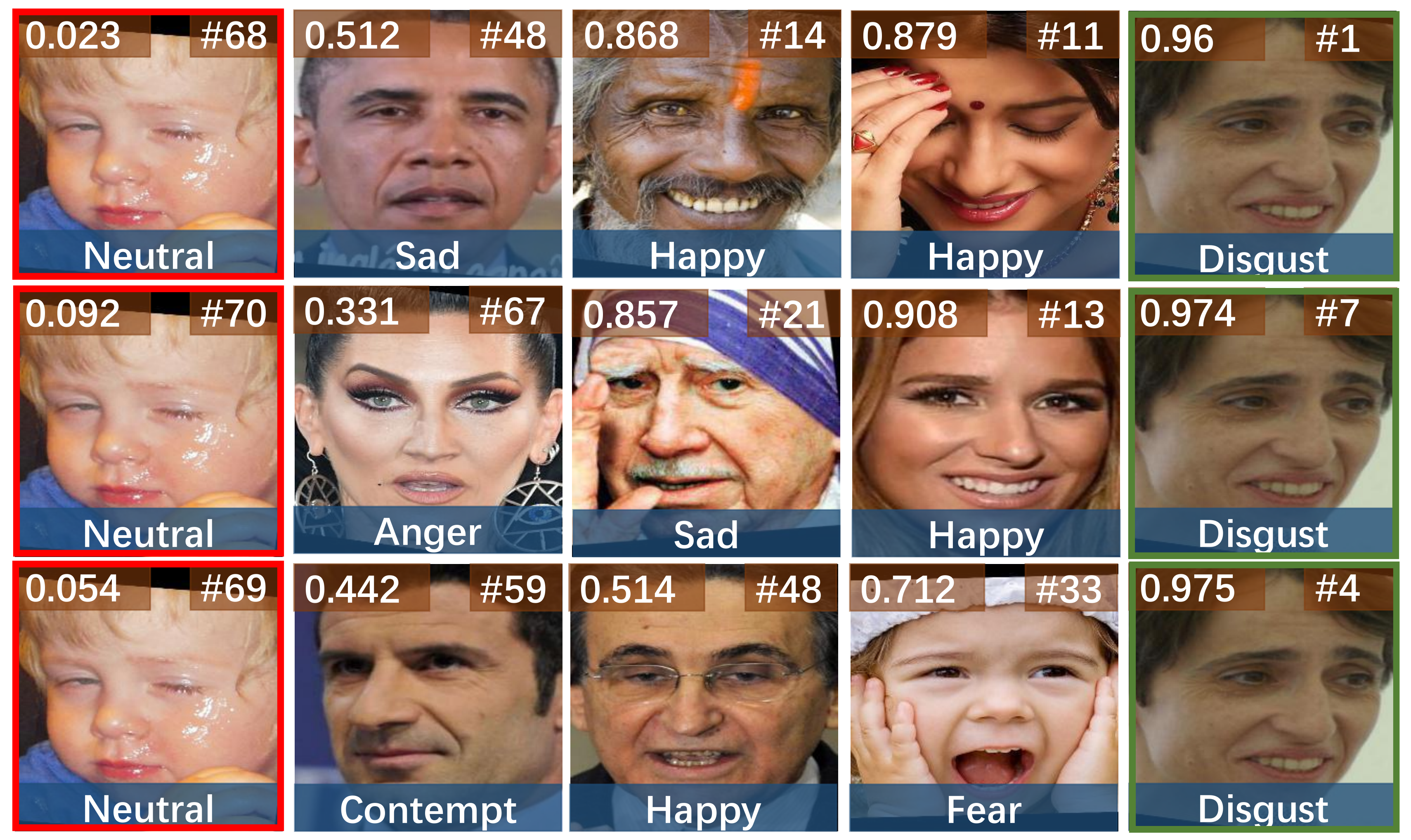}
    \vspace{-0.1cm}
    \caption{From top to bottom: images from three different batches with their annotations. Red (Green) bounding box denotes bad (good) anchor image.
    The upper left (right) corner of each picture is tagged with its confidence score (rank) in the batch. The estimated score is robust and consistent with intuition. Best viewed in color. Zoom in for better view.}
    \label{fig:score}
\end{figure}

In Fig.~\ref{fig:score}, we choose two typical anchor images and report their results in three different batches. The confident samples are assigned with higher score, while the ambiguous ones are the opposite. Furthermore, both the scores and ranks of anchor images are consistent within three different batches. It shows the robustness of our pairwise uncertainty estimation module. \textit{More analyses are provided in the supplementary material.}

\begin{table*}[t]	
	\centering
		\caption{Comparison with the state-of-the-art results. Res denotes ResNet. $^+$ denotes both AffectNet and RAF-DB are used as the training set. $^*$ means using extra distribution data instead of single category annotation. $^\dagger$ denotes the method is trained and tested with 7 classes on AffectNet.}
		\vspace{-0.6em}
	\begin{subtable}[t]{.25\linewidth} 
		\centering
		\caption{Comparison on AffectNet}\label{table:affectnet}
		\resizebox{.86\linewidth}{!}{%
		 \begin{tabular}{cc}
             \thickhline
             Method & Acc. \\
             \hline
             Upsample~\cite{DBLP:journals/affectnet} & 47.01 \\
             IPA2LT$^+$~\cite{DBLP:conf/eccv/IPA2LT} &  55.71\\
             RAN~\cite{DBLP:journals/affectnet} & 59.50 \\
             CAKE$^\dagger$~\cite{DBLP:conf/bmvc/cake} & 61.70 \\
             SCN~\cite{DBLP:conf/cvpr/SCN} & 60.23 \\
             \hline
             Ours(Res-18) & \textbf{62.84} \\
             Ours(Res-50IBN) & \textbf{63.11} \\
         \thickhline
         \end{tabular}%
         }%
	\end{subtable}%
	\begin{subtable}[t]{.25\linewidth}
		\centering
		\caption{Comparison on RAF-DB}\label{table:raf}
		\resizebox{.88\linewidth}{!}{%
		 \begin{tabular}{cc}
             \thickhline
             Method & Acc. \\
             \hline
             gaCNN~\cite{DBLP:journals/tip/gaCNN} & 85.07 \\
             LDL-ALSG$^+$~\cite{DBLP:conf/cvpr/LDL-ALSG} & 85.53 \\
             IPA2LT$^+$~\cite{DBLP:conf/eccv/IPA2LT} & 86.77 \\
             SCN~\cite{DBLP:conf/cvpr/SCN} & 87.03 \\
             SCN$^+$~\cite{DBLP:conf/cvpr/SCN} & 88.14 \\
             \hline
             Ours(Res-18) & \textbf{88.76} \\
             Ours(Res-50IBN) & \textbf{89.42} \\
         \thickhline
         \end{tabular}%
         }%
	\end{subtable}%
	\begin{subtable}[t]{.25\linewidth}
		\centering
		\caption{Comparison on FERPlus}\label{table:ferplus}
		\resizebox{.86\linewidth}{!}{%
		 \begin{tabular}{cc}
             \thickhline
             Method & Acc. \\
             \hline
             PLD$^*$~\cite{DBLP:conf/icmi/ferplus} & 85.10 \\
             Res+VGG~\cite{resvgg} & 87.40 \\
             SCN & 88.01 \\
             SeNet50$^*$~\cite{SeNet50} & 88.80 \\
             RAN~\cite{DBLP:journals/tip/RAN} & 88.55 \\
             \hline
             Ours(Res-18) & \textbf{88.64} \\
             Ours(Res-50IBN) & \textbf{89.51} \\
         \thickhline
         \end{tabular}%
         }%
	\end{subtable}%
	\begin{subtable}[t]{.25\linewidth}
		\centering
		\caption{Comparison on SFEW}\label{table:sfew}
		\resizebox{.86\linewidth}{!}{%
		 \begin{tabular}{cc}
             \thickhline
             Method & Acc. \\
             \hline
             IdentityCNN~\cite{DBLP:conf/icip/identitycnn} & 50.98 \\
             Island loss~\cite{DBLP:conf/fgr/island} & 52.52 \\
             Incept-ResV1~\cite{DBLP:conf/cvpr/cov_pool} & 51.90 \\
             MultiCNNs~\cite{DBLP:conf/icmi/multicnns} & 55.96 \\
             RAN~\cite{DBLP:journals/tip/RAN} & 56.40 \\
             \hline
             Ours(Res-18) & \textbf{57.12} \\
             Ours(Res-50IBN) & \textbf{58.34} \\
         \thickhline
         \end{tabular}%
         }%
	\end{subtable}%
	\label{table:state-of-the-art}
	\vspace{-0.5em}
\end{table*}

\subsection{Ablation Study}\label{sec:ablation}
We conduct extensive ablation studies on AffectNet, as it is the largest dataset. \textit{Some of them are provided in the supplementary material, due to the page limitation.}

\textbf{Mining latent distribution.} Quantitative and qualitative experiments on AffectNet are conducted to analyze the way of mining latent distribution. For the former, given a batch, we train each auxiliary branch with all the samples, where the ($C-1$)-class classifier is switched to $C$-class. To make the latent distribution, their predictions are averaged to increase the robustness. For simplification, we denote latent distribution mined in this way as LD-A, while the original in DUME as LD-N.

As shown in Table~\ref{tab:toy}, LD-N guides the target branch better. Because it can reflect more discriminative latent truth. \textit{More analyses are provided in supplementary material.}

\begin{table}
\vspace{-1em}
\centering
\caption{Ablation study of ways to mine latent distribution.}
\vspace{-0.8em}
    \resizebox{.68\columnwidth}{!}{
    \begin{centering}
        \begin{tabular}{c|c|c|c}
            \thickhline
            Methods &  Baseline &  LD-A & LD-N\\
            \hline
            Acc.~(\%)& 58.85 & 60.03 & 61.32\\
            \thickhline
            \end{tabular} %
    \end{centering}
    }
\label{tab:toy}
\end{table}
\begin{figure}[t]
  \begin{subfigure}[b]{0.49\columnwidth}
    \includegraphics[width=1\columnwidth]{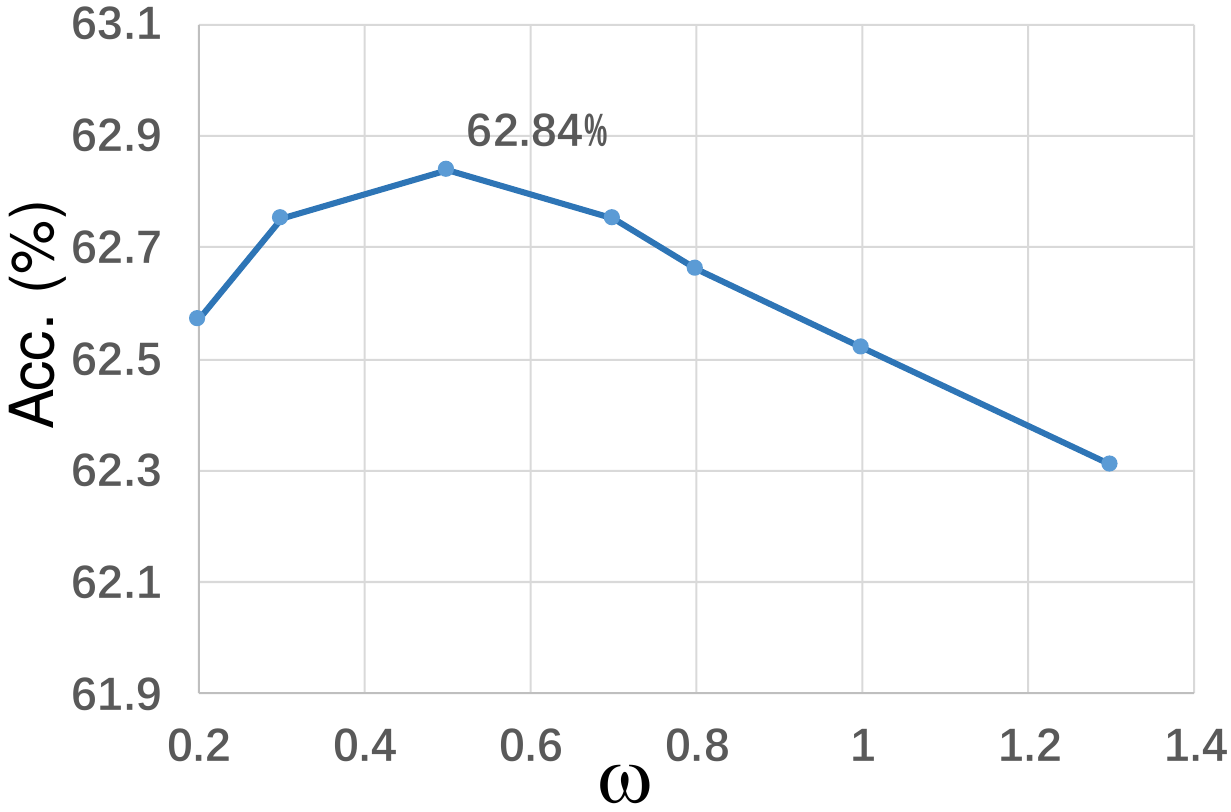}
    \caption{}
    \label{fig:ablation_w}
  \end{subfigure}
  \begin{subfigure}[b]{0.49\columnwidth}
    \includegraphics[width=1\columnwidth]{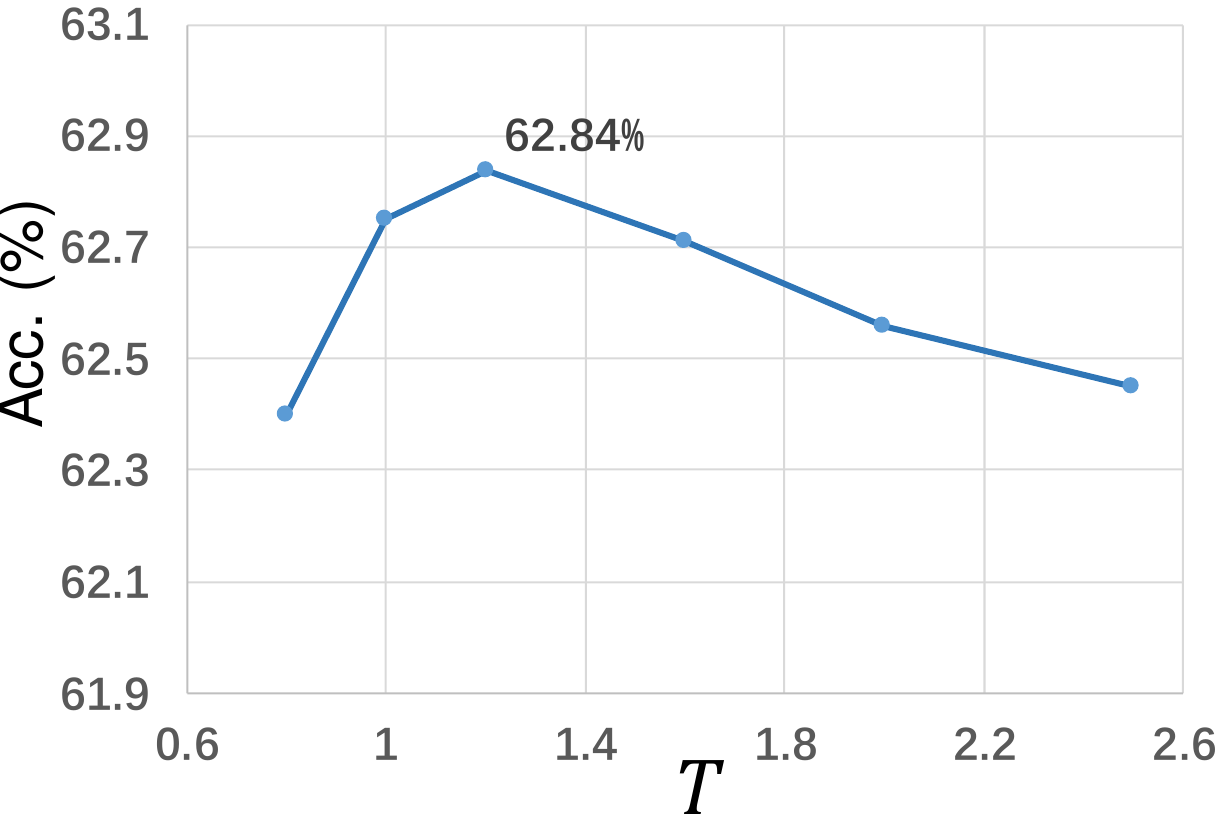}
    \caption{}
    \label{fig:ablation_T}
  \end{subfigure}
  \vspace{-0.5em}
  \caption{(a) The accuracy (\%) with different $\omega$. (b) The accuracy (\%) with different $T$.}
  \label{fig:ablation_wt}
\end{figure}

\textbf{Trade-off Weight $\omega$.}
$\omega$ balances the learning of target branch between $\widetilde{\boldsymbol{y}}_x$ and annotation. Fig.~\ref{fig:ablation_wt} shows that too small $\omega$ causes trouble for target branch to learn $\widetilde{\boldsymbol{y}}_x$. When $\omega$ is too large, it is hard for uncertain estimation module to adjust learning focus, as the sensitivity to $\widetilde{\boldsymbol{y}}_x$ is enlarged.

\textbf{Sharpen Temperature $T$.}
$T$ provides the flexibility to slightly modify the entropy of $\widetilde{\boldsymbol{y}}_x$. Fig.~\ref{fig:ablation_wt} shows the effect with different $T$. When $T < 1$, the distribution becomes steep quickly, damaging the fine-grained label information. Using $T>1$ flattens $\widetilde{\boldsymbol{y}}_x$, relieving model's sensitivity to incorrect predictions. Yet, the performance will be degraded if $T$ is too large, as the pattern of $\widetilde{\boldsymbol{y}}_x$ is suppressed.

\textbf{Epoch Threshold $\beta$.}
The first $\beta$-th epoch is dedicated to pretraining the auxiliary branches in prior, to make them provide stable latent distribution. After the $\beta$-th epoch, attention is paid more on optimizing the target branch. Table~\ref{tab:beta} shows the accuracy with different $\beta$.

\begin{figure}[t]
    \centering
    \vspace{-0.8em}
    \includegraphics[width=0.74\linewidth]{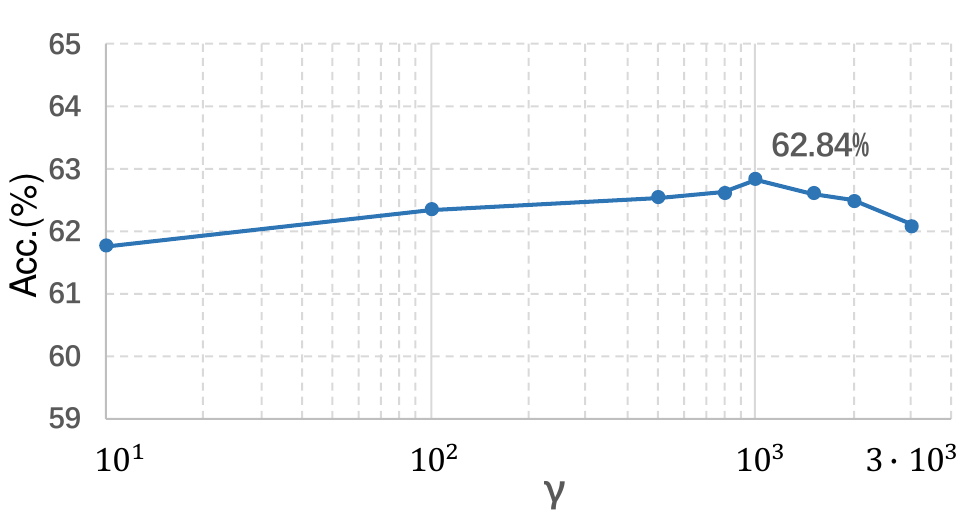}
    \vspace{-1em}
    \caption{Accuracy(\%) sensitivity to $\gamma$.}
    \label{fig:gamma}
\end{figure}

\textbf{Similarity Preserving factor $\gamma$.}
We generalized the similarity preserving to the context of multi-branch architecture. $\gamma$ adjusts the contribution ratio of the mechanism. Fig.~\ref{fig:gamma} reflects the performance of model with different $\gamma$.

\begin{table}
\vspace{-1em}
\centering
\caption{The accuracy (\%) with different $\beta$.}
\vspace{-0.8em}
    \resizebox{.88\columnwidth}{!}{
    \begin{centering}
        \begin{tabular}{c|c|c|c|c|c}
            \thickhline
            $\beta$ &  2 &  3 & 6 & 10 & 14 \\
            \hline
            Acc.~(\%)& 62.28 & 62.54 & 62.84 & 62.50 & 62.41\\
            \thickhline
            \end{tabular} %
    \end{centering}
    }
\label{tab:beta}
\end{table}

\vspace{-1pt}
\section{Conclusion}
\vspace{-1pt}
In order to address the ambiguity problem in FER, we propose DMUE, with the design of latent distribution mining and pairwise uncertainty estimation. On one hand, the mined latent distribution describes the ambiguous instance in a fine-grained way to guide the model. On the other hand, pairwise relationships between samples are fully exploited to estimate the ambiguity degree. Our framework imposes no extra burden on inference, and can be flexibly integrated with the existing network architectures. Experiments on popular benchmarks and synthetic ambiguous datasets show the effectiveness of DMUE.

\section*{Acknowledgements}
This work is supported by the National Key R\&D Program of China under Grant No. 2020AAA0103800 and by the National Natural Science Foundation of China under Grant U1936202 and 62071216. This work is mainly done at JD AI Research.

\vspace{2pt}
\section{Appendix}

\subsection{The Generalized Similarity Preserving Loss}
In this section, we give the generalized formulation of $L_{sp}$, which is defined as the following in the main text:
\begin{equation}
    \centering
    {L_{sp}} = MSP(G_{aux}^1,\cdots,G_{aux}^C, G_{tar}),
    \label{eq:ori_sp}
\end{equation}
where $G_{aux}^i\in\mathbb{R}^{N_i\times{N_i}}$ and $G_{tar}\in\mathbb{R}^{N\times{N}}$ denote the similarity matrices calculated by the auxiliary and target branches, respectively. $N$ is the batch size, and $N_i$ is the number of images that are not annotated to the $i$-th class in the batch. For coding simplicity, given an image batch, we define $\boldsymbol{A}_{tar}\in\mathbb{R}^{N\times{N}}$ and $\boldsymbol{A}_{aux}^{i}\in\mathbb{R}^{N\times{N}}$ ($i\in\{1,2,\cdots,C\}$), and the $j$-th row $\boldsymbol{a}_{tar_j}$ of $\boldsymbol{A}_{tar}$ and $\boldsymbol{a}_{aux_j}^i$ of $\boldsymbol{A}_{aux}^{i}$ are denoted as:
\begin{equation}
    \centering
        {\boldsymbol{a}_{tar_j}} =  {\frac{{ {{\boldsymbol{f}_{tar_j}}\cdot{\boldsymbol{f}_{tar}^{T}}} }}{{\left\| {{{\boldsymbol{f}_{tar_j}}\cdot{\boldsymbol{f}_{tar}^{T}}}} \right\|_2}},}
\end{equation}
\begin{equation}
    \centering
        {\boldsymbol{a}_{aux_j}^i} =  {\frac{{ {{\boldsymbol{f}_{aux_j}^i}\cdot{\boldsymbol{f}{_{aux}^{i}}^{T}}} }}{{\left\| {{\boldsymbol{f}_{aux_j}^i}\cdot{\boldsymbol{f}{_{aux}^{i}}^{T}}} \right\|_2}},}
\end{equation}
where $\boldsymbol{f}_{tar}\in\mathbb{R}^{N\times{d}}$ and $\boldsymbol{f}{_{aux}^{i}}\in\mathbb{R}^{N\times{d}}$ are the semantic features in the target branch and the auxiliary branch $i$, respectively, $\boldsymbol{f}_{tar_j}\in\mathbb{R}^{1\times{d}}$ and $\boldsymbol{f}_{aux_j}^i\in\mathbb{R}^{1\times{d}}$ are the $j$-th row of $\boldsymbol{f}_{tar}$ and $\boldsymbol{f}{_{aux}^{i}}$, respectively, and $d$ is the feature dimension. Then, we implement $L_{sp}$ by masking $\boldsymbol{A}_{tar}$ and $\boldsymbol{A}_{aux}^{i}$, which $L_{sp}$ can be rewritten as:
\begin{equation}
\label{eq:new_sp}
    \centering
    {L_{sp}} = \frac{1}{C}\sum\limits_{j = 1}^C {\frac{1}{{N_i^2}}} \left\|
    \boldsymbol{M}^i\ast{\boldsymbol{A}_{tar} - \boldsymbol{M}^i\ast\boldsymbol{A}_{aux}^i} \right\|_F^2,
\end{equation}
where $\ast$ is the element-wise product. The $q$-th row and $p$-th column element $m_{q,p}^i$ of $\boldsymbol{M}^i{\in\mathbb{R}^{N\times{N}}}$ is defined as:
\begin{equation}
\label{mask}
m_{q,p}^i = \left\{ {\begin{array}{*{20}{c}}
0&{y_p=i\ or\ y_q=i}\\
1&{Others}
\end{array}} \right.,
\end{equation}
where $y_p$ and $y_q$ denote the annotations of the $p$-th and $q$-th images in the batch, respectively ($y_p, y_q\in\{1,\cdots,C\}$). $L_{sp}$ is easy to be implemented by a few lines of code\footnote{Our source code and pre-trained models will be released.}.

\begin{figure}[t]
    \centering
    \vspace{-0.5em}
    \includegraphics[width=0.9\linewidth]{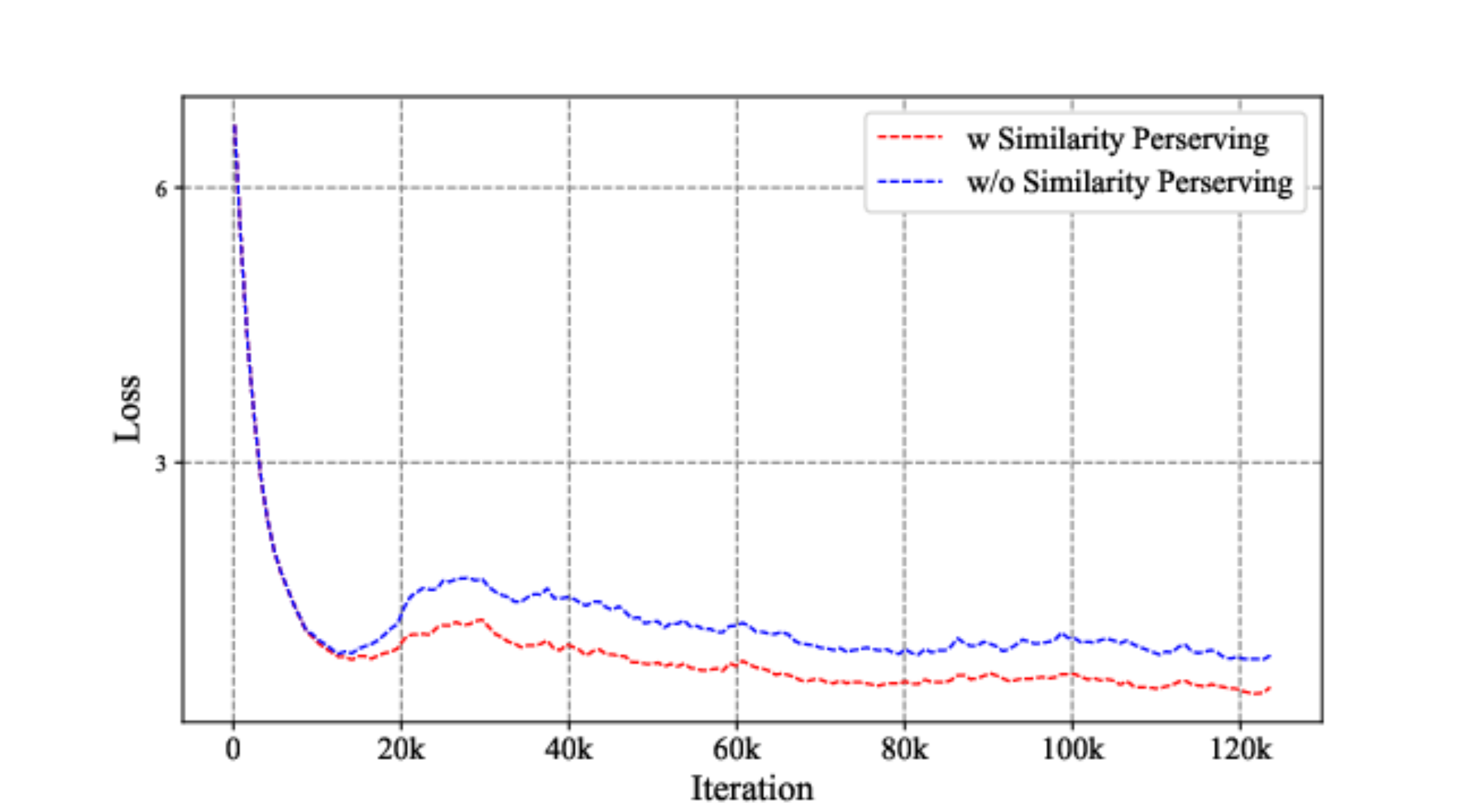}
    \vspace{-1em}
    \caption{The loss curves of training with $L_{sp}$ (the red curve) and without $L_{sp}$ (the blue curve) along iterations. Experiments are conducted on AffectNet with ResNet-18 as the backbone architecture.}
    \label{fig:sp_loss}
\end{figure}

\textbf{Benefits of $L_{sp}$.} The non-zero elements in ${\boldsymbol{A}}_{tar}$ and ${\boldsymbol{A}}_{aux}^i$ represent the predicted similarity values of image pairs. With the constraint of $L_{sp}$, all the branches are regularized to predict consistent similarity value for an image pair. As shown in Fig.~\ref{fig:sp_loss}, similarity preserving makes the training more stable. The loss value rises up a little around the $20k$-th iteration step because the ramp functions gradually assign larger weight for $L_{wce}^{target}$ to train the target branch.

\subsection{Evaluation of Synthetic Ambiguity}
\begin{figure*}[b]
    \centering
    \includegraphics[width=0.86\linewidth]{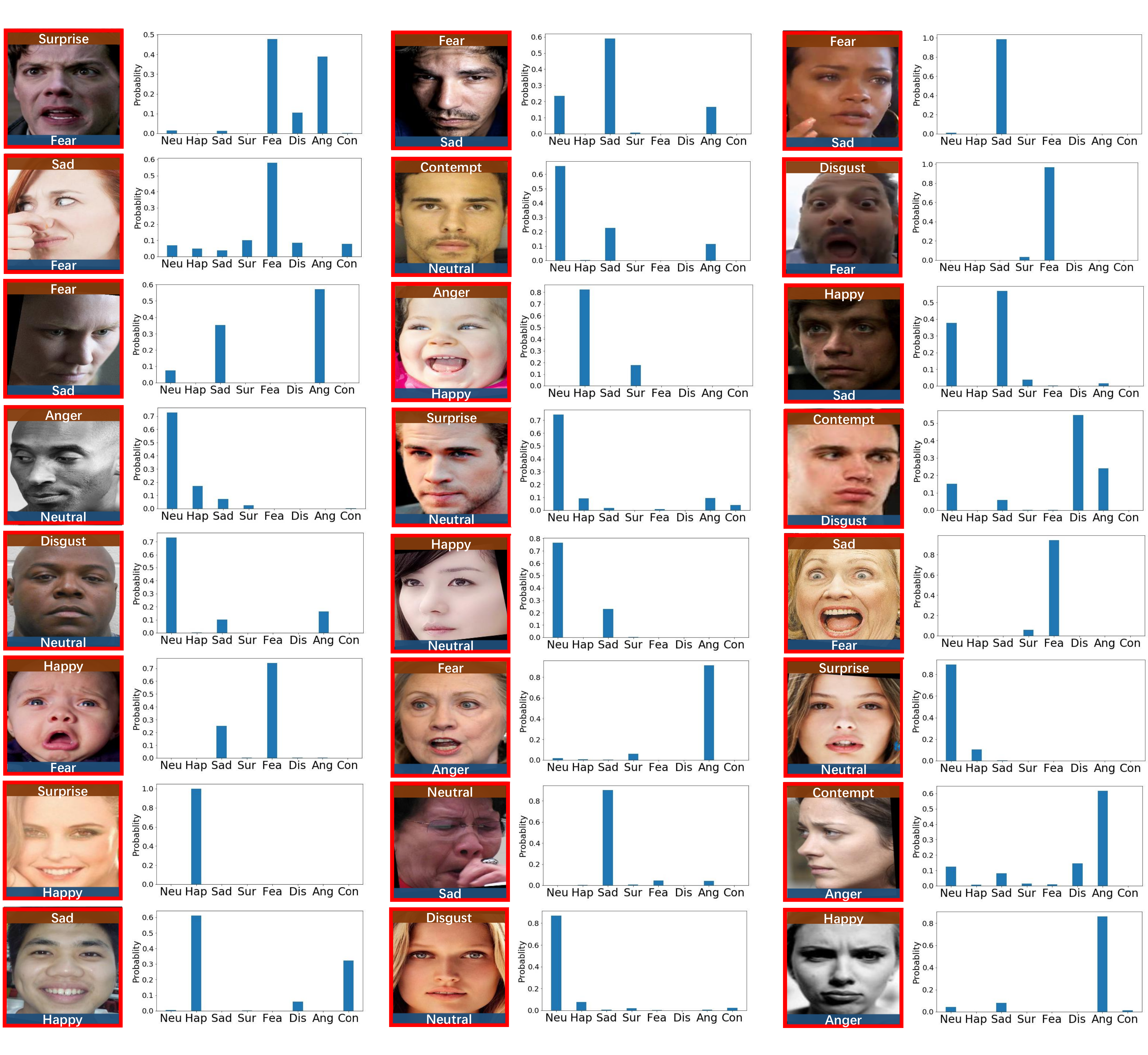}
    \caption{DMUE yields the latent truth for noisy samples. The bottom of each image is tagged by its original annotation. The top of each image is tagged by manually flipped noisy label. DMUE is adopted to train network on synthetic noisy datasets. We visualize the mined latent distribution for synthetic noisy samples at the right of each image. The mined latent distribution is in line with the human subjective perception, where the most possible class reflected by latent distribution is corresponding to the original annotation. (Neu=Neutral, Hap=Happy, Sad=Sad, Sur=Surprise, Fea=Fear, Dis=Disgust, Ang=Anger, Con=Contempt)}
    \label{fig:mislabelled}
    \vspace{-0.5em}
\end{figure*}

To better demonstrate the superiority of latent distribution mining, we conduct qualitative analyses on synthetic mislabelled samples. Specifically, a portion of training samples are randomly chosen of which the labels are flipped to other categories. Then, we use DMUE to train the network on the synthetic mislabelled data and visualize the mined latent distributions in Fig.~\ref{fig:mislabelled}. We can observe that the mined latent distribution is able to correct the noisy annotation well. Taking the last image in Fig.~\ref{fig:mislabelled} as an example, although its true label \textit{Anger} is flipped to \textit{Happy}, the latent distribution well reflects its true class \textit{Anger}. Moreover, the latent distribution also has the capacity to reflect the second possible class for compound expressions. For the first image, the latent distribution reflects its second possible class \textit{Anger}, which is in line with the subjective perception. By imposing the latent distribution as the additional supervision, DMUE effectively utilizes the semantic features of samples.

\subsection{Ablation Study}
\textbf{Different Backbone Networks.} As described in the main text, DMUE is independent to the backbone architectures. We further apply DMUE to ShuffleNetV1~\cite{DBLP:conf/cvpr/shufflenet} and MobileNetV2~\cite{DBLP:conf/cvpr/mobilenet} to demonstrate the universality of DMUE, where the last stage of ShuffleNetV1 and the last two stages of MobileNetV2 are separated for latent distribution mining. The results on AffectNet and RAF-DB are presented in Table~\ref{tab:msra_different}. We observe that DMUE can stably improve the performance of all the architectures, including ShuffleNetV1, MobileNetV2, ResNet-18 and ResNet-50IBN, by an average of 4.30\% and 2.47\% on AffectNet and RAF-DB, respectively. In addition, the ResNet50-IBN achieves the best record among these architectures because of the large number of parameters and the IBN module. Furthermore, we report the results of training DMUE from scratch on AffectNet and RAF-DB in Table~\ref{tab:no_msra_different}. Similar observations can also be found without using the pre-trained model on MS-Celeb-1M.

\begin{table}[t]
\vspace{-0.5em}
\begin{center}
\resizebox{\columnwidth}{!}{ 
\begin{tabular}{c|c|c|c}
\thickhline
Backbone Architecture & DMUE & AffectNet & RAF-DB\\
\thickhline
ShuffleNetV1 (group=3;2.0$\times$) & - & 56.51 & 86.20\\
ShuffleNetV1 (group=3;2.0$\times$) & \checkmark & \textbf{60.87} & \textbf{88.73}\\
MobileNetV2 & - & 57.65 & 86.01 \\
MobileNetV2 & \checkmark & \textbf{62.34} & \textbf{87.97} \\
ResNet-18 & - & 58.85 & 86.33\\
ResNet-18 & \checkmark & \textbf{62.84} & \textbf{88.76}\\
ResNet50-IBN & - & 58.94 & 86.57\\
ResNet50-IBN & \checkmark & \textbf{63.11} & \textbf{89.51}\\ \thickhline
\end{tabular}} %
\end{center}
\caption{Accuracy ($\%$) on RAF-DB and AffectNet with pre-training on MS-Celeb-1M.}
\label{tab:msra_different}
\vspace{-1em}
\end{table}
\begin{table}[t]
\vspace{-0.5em}
\begin{center}
\resizebox{\columnwidth}{!}{ 
\begin{tabular}{c|c|c|c}
\thickhline
Backbone Architecture & DMUE & AffectNet & RAF-DB\\
\thickhline
ShuffleNetV1 (group=3;2.0$\times$) & - & 55.02 & 85.65\\
ShuffleNetV1 (group=3;2.0$\times$) & \checkmark & \textbf{59.67} & \textbf{88.10}\\
MobileNetV2 & - & 54.94 & 85.44 \\
MobileNetV2 & \checkmark & \textbf{60.43} & \textbf{88.15} \\
ResNet-18 & - & 55.22 & 86.01\\
ResNet-18 & \checkmark & \textbf{61.22} & \textbf{88.33}\\
ResNet50-IBN & - & 55.52 & 85.67\\
ResNet50-IBN & \checkmark & \textbf{60.54} & \textbf{88.94}\\ \thickhline
\end{tabular}} %
\end{center}
\caption{Accuracy ($\%$) on RAF-DB and AffectNet without pre-training on MS-Celeb-1M.}
\label{tab:no_msra_different}
\vspace{-1em}
\end{table}

\begin{figure*}
    \centering
    \includegraphics[width=0.86\linewidth]{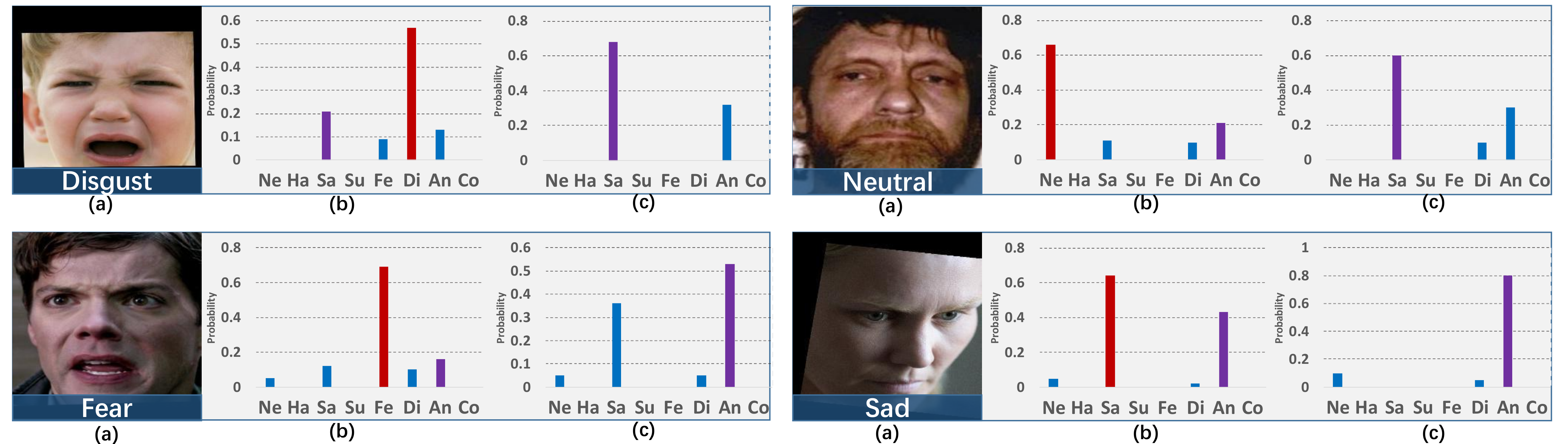}
    \caption{Qualitative comparison between LD-A and LD-N. The red bar denotes the positive class predicted in LD-A. The purple bar denotes the second possible class predicted in LD-A and LD-N for ambiguous images. (a) Images tagged by their original annotation. (b) The LD-A can reflect the visual feature to a certain extent, yet the images' possibility distribution among its negative classes is not discriminative. (c) The LD-N that we used in DMUE, describes an image on its negative classes discriminatively. (Ne=Neutral, Ha=Happy, Sa=Sad, Su=Surprise, Fe=Fear, Di=Disgust, An=Anger, Co=Contempt)}
    \label{fig:abandon}
    \vspace{-0.5em}
\end{figure*}

\textbf{Mining latent distribution.}
In the main text Table 4, we describe the quantitative comparison aiming at investigating which way to mine latent distribution is better. In the quantitative comparison, we train the auxiliary branches with the whole image batch, and their predictions for each image are averaged, denoted as LD-A. As shown in Fig.~\ref{fig:abandon}, LD-A reflects the visual feature of images to some extent. But the second and the third possible class of a compound expression is not discriminative in LD-A.

In Fig~\ref{fig:toy}, we provide a toy example of latent label space some intermediate iterations. The latent distribution is completely random at the beginning. It gradually reflects the visual feature of the sample during iterations.

\subsection{More Results of User Study}
\begin{figure*}
    \centering
    \includegraphics[width=0.86\linewidth]{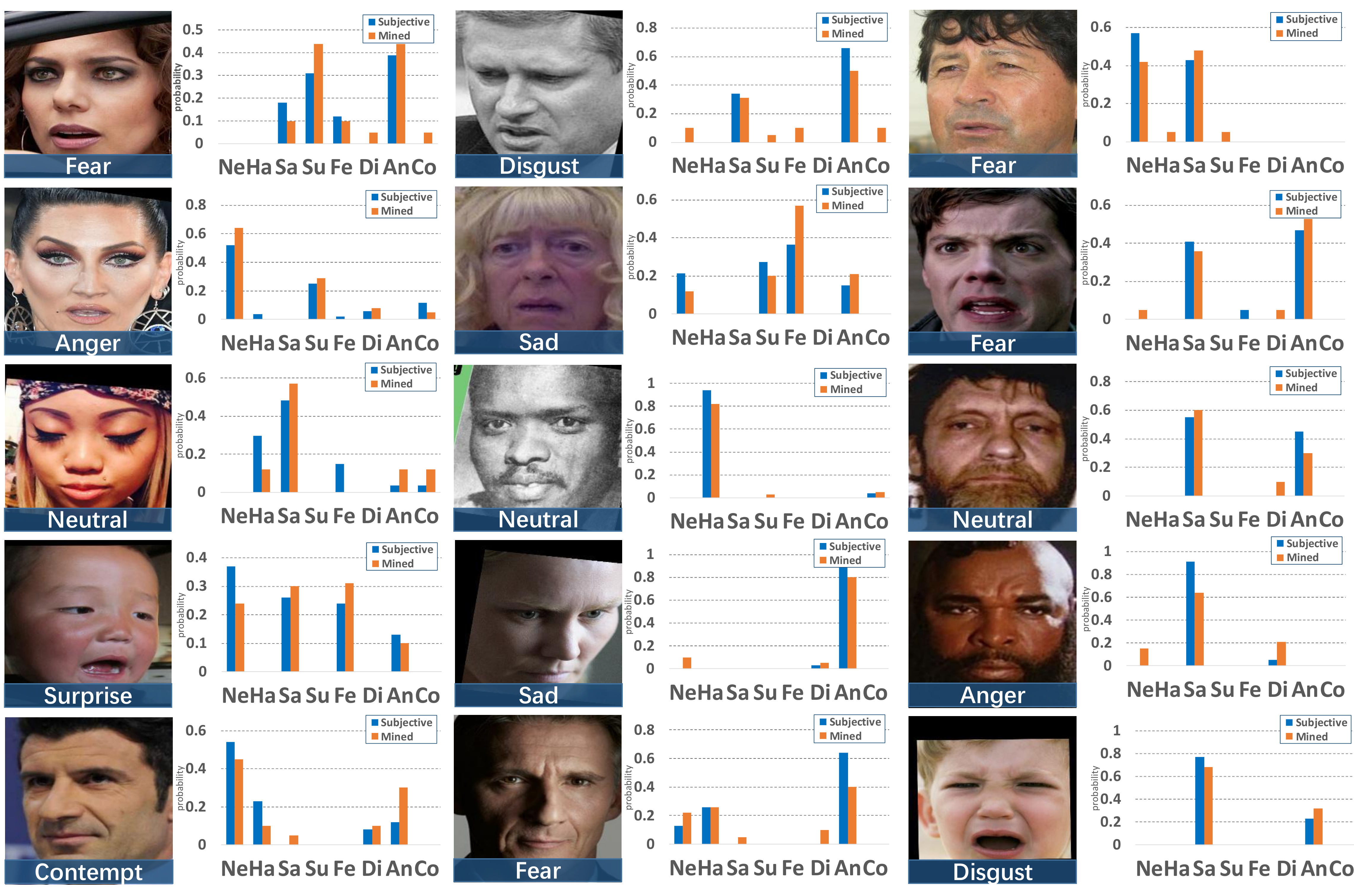}
    \caption{More visualizations of the mined latent distributions and subjective survey results. The age of 50 volunteers range from 17 to 51. Each image is tagged with its annotation. The orange bar denotes the mined latent distribution and the blue bar denotes the subjective survey results. To process the votes from volunteers, we set the number of votes on each sample's positive class as zero. Then we normalize the results, which reflect the probability that image belonging to each negative class. As we can see, the mined latent distribution is consistent with human intuition in general.
    (Ne=Neutral, Ha=Happy, Sa=Sad, Su=Surprise, Fe=Fear, Di=Disgust, An=Anger, Co=Contempt)}
    \label{fig:more_label_res}
    \vspace{-0.5em}
\end{figure*}

As described in the main text, we pick 20 images from FER datasets and have them labelled by 50 volunteers. We provide more visualization results of the mined latent distribution and the perception from volunteers in Fig.~\ref{fig:more_label_res}. Accordingly, we draw the following conclusions: (\textbf{1}) One inherent property of facial expression is that compound facial expressions may exist. It is easy for volunteers to have disagreements with the exact type of images whose annotations in dataset are \textit{Fear}, \textit{Disgust}, \textit{Sad} and \textit{Anger}. One reason may be that folds in the region of eyebrow are often involved in those easily confused expressions. Thus, they may share some common visual features, making it hard to define the exact expression type from a static image. (\textbf{2}) The main goal of latent distribution is to provide reasonable guidance to the target branch, rather than finding the exact label distribution of a facial expression image. Thus, we utilize the $L_2$ loss to minimize the deviation because it is bounded and less sensitive to the incorrect prediction.

\textbf{Why auxiliary branches work?} Based on the conclusions above, we find the one-hot label is hard to represent the visual features of expressions. The annotation of face expression is subjective and difficult, because different expressions naturally entangle each other in the visual space. Auxiliary branches are proposed to disentangle such connections. Each auxiliary branch is a classifier that maps images to their latent classes. By doing so, we disentangle the ambiguity in the label space.

\par\noindent
\begin{figure}[t]
\vspace{0em}
    \centering
    \includegraphics[width=1\linewidth]{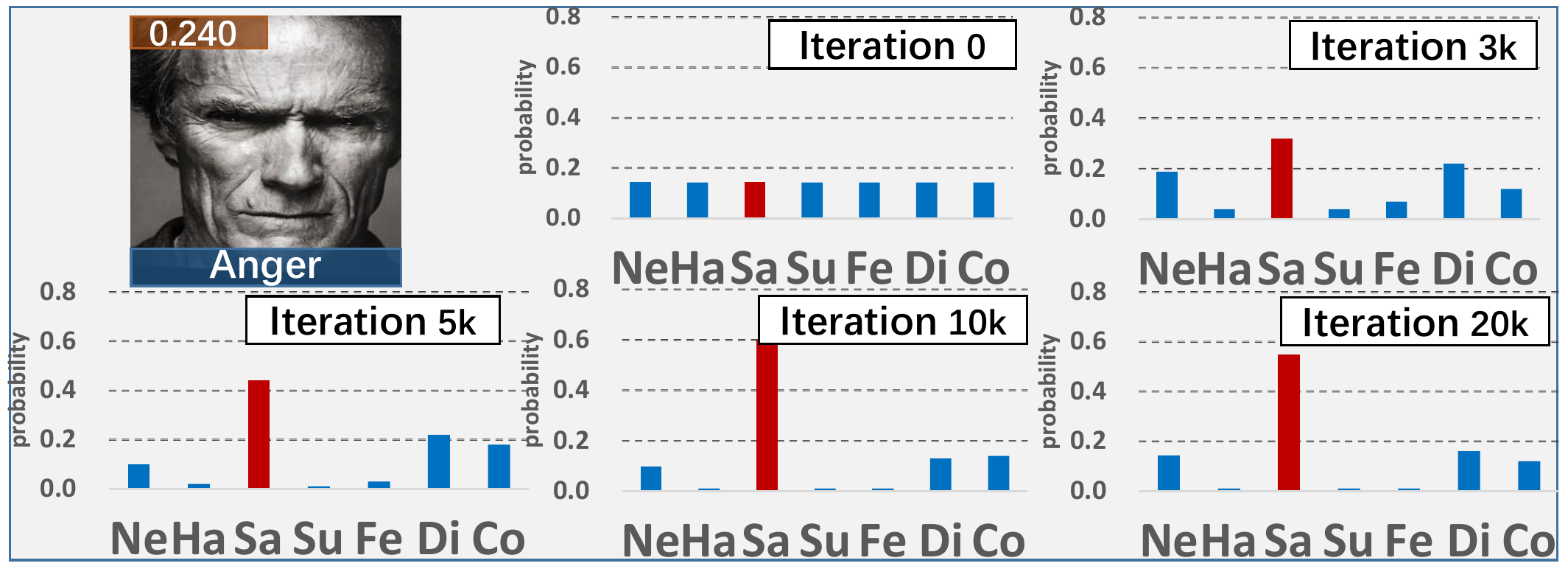}
    \vspace{-1.5em}
    \caption{The iteratively updated latent distribution of the sample from Fig. 4 in the main text. Best viewed in color. Zoom in for better view.
    }
    \label{fig:toy}
    \vspace{-0.5em}
\end{figure}
\par\noindent

\subsection{More Results and Mathematical Reason for the Uncertainty Estimation}
Assume a class center feature $\boldsymbol{c}_i$ of the $i$-th class, the angle between a $i$-th class sample's feature $\boldsymbol{x}$ and $\boldsymbol{c}_i$ is $\theta$, that $\theta=acos(\left\langle\boldsymbol{x},\boldsymbol{c}_i\right\rangle)$. Without losing generality, assume $\theta\sim\mathcal{N}(0,\sigma^2)$ in [$-\pi,\pi$] (as the ambiguity increases, the number of samples decreases). Given a sample $\boldsymbol{a}$ with semantic feature $\boldsymbol{f}$, that  $\left\langle\boldsymbol{f},\boldsymbol{c}_i\right\rangle=cos\alpha$. We have:
\par\noindent
\begin{equation}
\centering
    {{S}_{a,i}}=\mathop E\limits_{\theta\sim\mathcal{N}(0,{\sigma ^2}),\theta  \in [ -\pi,\pi]} \{  < \boldsymbol{x},\boldsymbol{f} > \},
\end{equation}
where $\|\boldsymbol{c}_i\|=\|\boldsymbol{x}\|=\|\boldsymbol{f}\|=1$. We first study the equation in a special case where $\alpha=\frac{\pi}{2}$. We denote $\boldsymbol{x}^{\bot}$ as the projection from $\boldsymbol{x}$ to the linear subspace $\boldsymbol{W}$:
\par\noindent
\begin{equation}
    \centering
    \boldsymbol{x}=\boldsymbol{c}_i+\boldsymbol{x}^{\bot},
\end{equation}
$\mathbb{R}^{n}$ = $span\{\boldsymbol{c}_i\} \oplus W$, $\oplus$ is the direct sum. Let $\{\boldsymbol{z}_1,\cdots,\boldsymbol{z}_{n-1}\}$ is a set of basis of $\boldsymbol{W}$. We have $\boldsymbol{x}^{\bot}$ under uniform distribution as prior for Softmax or other angle-based loss, that is $E\{\left<\boldsymbol{x}^{\bot},\boldsymbol{z}_{k}\right>\}=0$, with $1\leq k\leq n-1$.
As $\alpha=\frac{\pi}{2}$ in this special case, $\boldsymbol{f}$ can be rewritten as:
\par\noindent
\begin{equation}
\centering
 \boldsymbol{f} = \sum\nolimits_k^{n-1} {\omega_{k}\boldsymbol{z}_k}.
\end{equation}
We have $\boldsymbol{f}\bot\boldsymbol{c}_i$, and have:
\par\noindent
\begin{equation}
\begin{aligned}
     {E\{<\boldsymbol{x}^{\bot},\boldsymbol{f}>\}}&=E\{\boldsymbol{x}^{\bot}\cdot\sum\nolimits_k^{n-1}{\omega_{k}\boldsymbol{z}_k}\}\\
     &=\sum\nolimits_k^{n-1}{\omega_{k}E\{\boldsymbol{x}^{\bot}\cdot{\boldsymbol{z}_k}}\}\\
     &=0,\\
     E\{<\boldsymbol{x}, \boldsymbol{f}>\} &= 0.
\end{aligned}
\label{eq:lemma}
\end{equation}
\par\noindent
Now, for the general case, we construct:
\par\noindent
\begin{equation}
\left\{
             \begin{array}{lr}
             \boldsymbol{f}_1= <\boldsymbol{f},\boldsymbol{c}_i>\boldsymbol{c_i}, &  \\
             \boldsymbol{f}_2=\boldsymbol{f}-\boldsymbol{f}_1, &
             \end{array}
\right.
\end{equation}
\par\noindent
\begin{equation}
\centering
< \boldsymbol{x}, \boldsymbol{f} >=<\boldsymbol{x}, \boldsymbol{f}_1>+<\boldsymbol{x}, \boldsymbol{f}_2>.
\end{equation}
We notice that $\boldsymbol{f}_2\bot\boldsymbol{c}_i$:
\par\noindent
\begin{equation}
\begin{aligned}
     < \boldsymbol{f}_2, \boldsymbol{c}_i > &= <\boldsymbol{f}, \boldsymbol{c}_i> - <\boldsymbol{f}_1, \boldsymbol{c}_i> \\
     &=<\boldsymbol{f}, \boldsymbol{c}_i> - <<\boldsymbol{f},\boldsymbol{c}_i>\boldsymbol{c_i}, \boldsymbol{c}_i> \\
     &=<\boldsymbol{f},\boldsymbol{c}_i> - <\boldsymbol{f},\boldsymbol{c}_i>\|\boldsymbol{c}_i\| \\
     &= 0.
\end{aligned}
\end{equation}
\par\noindent
From Eq.~\ref{eq:lemma}, we have $E\{\left<\boldsymbol{x}, \boldsymbol{f}_2\right>\}=0$, so we have:
\par\noindent
\begin{equation}
    \begin{aligned}
    {{S}_{a,i}} &=
         \mathop E\limits_{\theta\sim\mathcal{N}(0,{\sigma ^2}),\theta  \in [ -\pi,\pi]}\{<\boldsymbol{x}, \boldsymbol{f}>\} \\
         &= \mathop E\limits_{\theta\sim\mathcal{N}(0,{\sigma ^2}),\theta  \in [ -\pi,\pi]}\{<\boldsymbol{x}, \boldsymbol{f}_1>\} \\
         &= \mathop E\limits_{\theta\sim\mathcal{N}(0,{\sigma ^2}),\theta  \in [ -\pi,\pi]}\{<\boldsymbol{x}, <\boldsymbol{f},\boldsymbol{c}_i>\boldsymbol{c_i}>\} \\
         &=<\boldsymbol{f},\boldsymbol{c}_i>\mathop E\limits_{\theta\sim\mathcal{N}(0,{\sigma ^2}),\theta  \in [ -\pi,\pi]}\{<\boldsymbol{x},\boldsymbol{c_i}>\} \\
         &=\cos \alpha \mathop E\limits_{\theta\sim\mathcal{N}(0,{\sigma ^2}),\theta  \in [ -\pi,\pi]} \{ \cos \theta \}.
    \end{aligned}
\end{equation}

Obviously, if $\boldsymbol{a}$ is the $j$-th class sample mislabelled to the $i$-th class, then $|\alpha|$ is large, ${S}_{a,i}$ becomes small and ${S}_{a,j}$ becomes large, which is contrary to the concatenated label. Thus, we can estimate the uncertainty from $\boldsymbol{SV}_a$ as it carries ambiguity information.

\begin{figure*}
    \centering
    \includegraphics[width=0.94\linewidth]{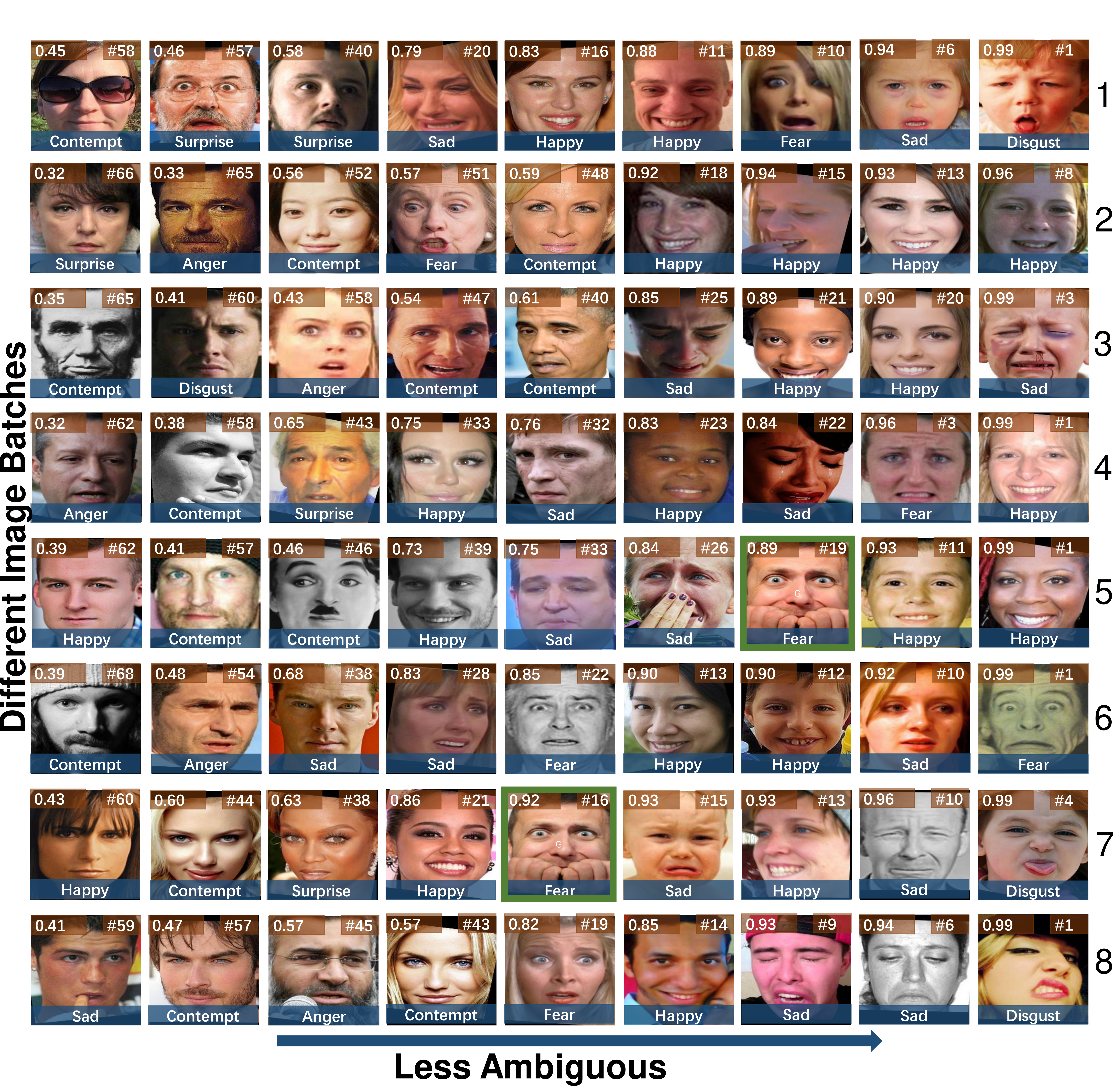}
    \caption{More visualization results of the estimated confidence score. From top to bottom, each row presents images from the same batch. The bottom of each image is tagged with its annotation from the dataset. The upper left of each image is tagged with its estimated uncertainty score. Lower scores are assigned to those more ambiguous images. The upper right of each image is tagged with its confidence rank in the batch. From left to right, we present images with their confidence scores in an ascending order. Images near the right side of the figure are less ambiguous, while images near the left side are in the opposite. In general, we observe that the estimated uncertainty score is in line with the subjective perception. Moreover, we insert an anchor image which is annotated to \textit{Fear} in two different batches (the green bounding box in the $5$-th and $7$-th row). The uncertainty estimation module predicts consistent confidence score for this anchor image, which indicates the stability of our uncertainty estimation module.}
    \label{fig:more_score}
    \vspace{-0.5em}
\end{figure*}

We present more visualization results of the estimated uncertainty score in Fig.~\ref{fig:more_score}, where lower scores mean more ambiguous images. It is obvious that the estimated uncertainty level is in line with the subjective perception. With the uncertainty estimation module, DMUE is able to suppress the adverse influence from ambiguous data, encouraging the network to utilize the semantic features and learn the latent distribution for the ambiguous image.

{\small
\bibliographystyle{ieee_fullname}
\bibliography{egbib}
}

\end{document}